\documentclass[conference]{IEEEtran}
\IEEEoverridecommandlockouts
\usepackage{amsmath,amssymb,amsfonts}
\usepackage{graphicx}

\usepackage{textcomp}
\usepackage{xcolor}
\usepackage{multirow}
\usepackage{float}
\usepackage{mathtools}
\usepackage{booktabs}
\usepackage{hyperref}

\usepackage{algpseudocode}
\algrenewcommand\algorithmicrequire{\textbf{Input:}}
\algrenewcommand\algorithmicensure{\textbf{Output:}}

\def\BibTeX{{\rm B\kern-.05em{\sc i\kern-.025em b}\kern-.08em
    T\kern-.1667em\lower.7ex\hbox{E}\kern-.125emX}}
    
\newcommand{\grad}{\triangledown}
\newcommand{\R}{\mathbb R}
\newcommand{\bx}{\mathbf{x}}
\newcommand{\bz}{\mathbf{z}}
\newcommand{\bb}{\mathbf{b}}
\newcommand{\bZ}{{\mathbf{Z}}}

\newcommand{\transname}{\textrm{Discrete CDT Transform}} 

\newcommand{\wtheta}{\mathbf{w}_\theta}
\newcommand{\mH}{\mathbb H}
\newcommand{\mG}{\mathbb G}
\newcommand{\mP}{\mathbb{P}}

\newcommand{\hogname}{\textrm{Discrete R-CDT transform}}
\newcommand{\HOG}{\F^*}
\newcommand{\F}{\mathcal F}

\newcommand{\wass}[2]{W_2(#1,#2)}

\newcommand{\transpace}{\big(\R^N\big)^{[0,\pi)}}

\DeclareMathAlphabet{\mathpzc}{OT1}{pzc}{m}{it}

\newcommand{\V}{\mathbb{V}}

\newcommand{\ltwog}{L^2([0,\pi), \R^N)}

\newcommand{\nImagePatch}{{I_k}}

\newcommand{\nDistriP}{{P_{\nabla I_k}}}

\newcommand{\nOmegaNPoints}{{\Omega^N_k}}

\newcommand{\nDistriPh}{{P^h_{\nabla I_{k}}}}
\newcommand{\nSetH}{{\mP_{\mathbb{H},k}}}

\newcommand{\nSetHNot}{{\mP_{\mathbb{H}_0,k}}}

\newcommand{\nOurMethod}{\text{Discrete R-CDT + NS}}

\begin{document}

\title{Local Sliced-Wasserstein Feature Sets for Illumination-invariant Face Recognition}


\author{\IEEEauthorblockN{Yan Zhuang\textsuperscript{1,2}, Shiying Li\textsuperscript{1,3}, Mohammad Shifat-E-Rabbi\textsuperscript{1,3}, Xuwang Yin\textsuperscript{1,2}\\
Abu Hasnat Mohammad Rubaiyat\textsuperscript{1,2}, Gustavo K. Rohde\textsuperscript{1,2,3}}
\\ Imaging and Data Science Laboratory\textsuperscript{1}\\ Department of Electrical and Computer Engineering\textsuperscript{2}, Department of Biomedical Engineering\textsuperscript{3} \\
University of Virginia
}
\maketitle 
\begin{abstract}
We present a new method for face recognition from digital images acquired under varying illumination conditions. The method is based on mathematical modeling of local gradient distributions using the Radon Cumulative Distribution Transform (R-CDT) \cite{kolouri2015radon}. We demonstrate that lighting variations cause certain types of deformations of local image gradient distributions which, when expressed in R-CDT domain, can be modeled as a subspace. Face recognition is then performed using a nearest subspace in R-CDT domain of local gradient distributions. Experiment results demonstrate the proposed method outperforms other alternatives in several face recognition tasks with challenging illumination conditions. Python code implementing the proposed method is available at \cite{software}, which is integrated as a part of the software package PyTransKit \cite{package}.
\end{abstract}


\section{Introduction}

\begin{figure}
    \centering
    \includegraphics[width=0.48\textwidth]{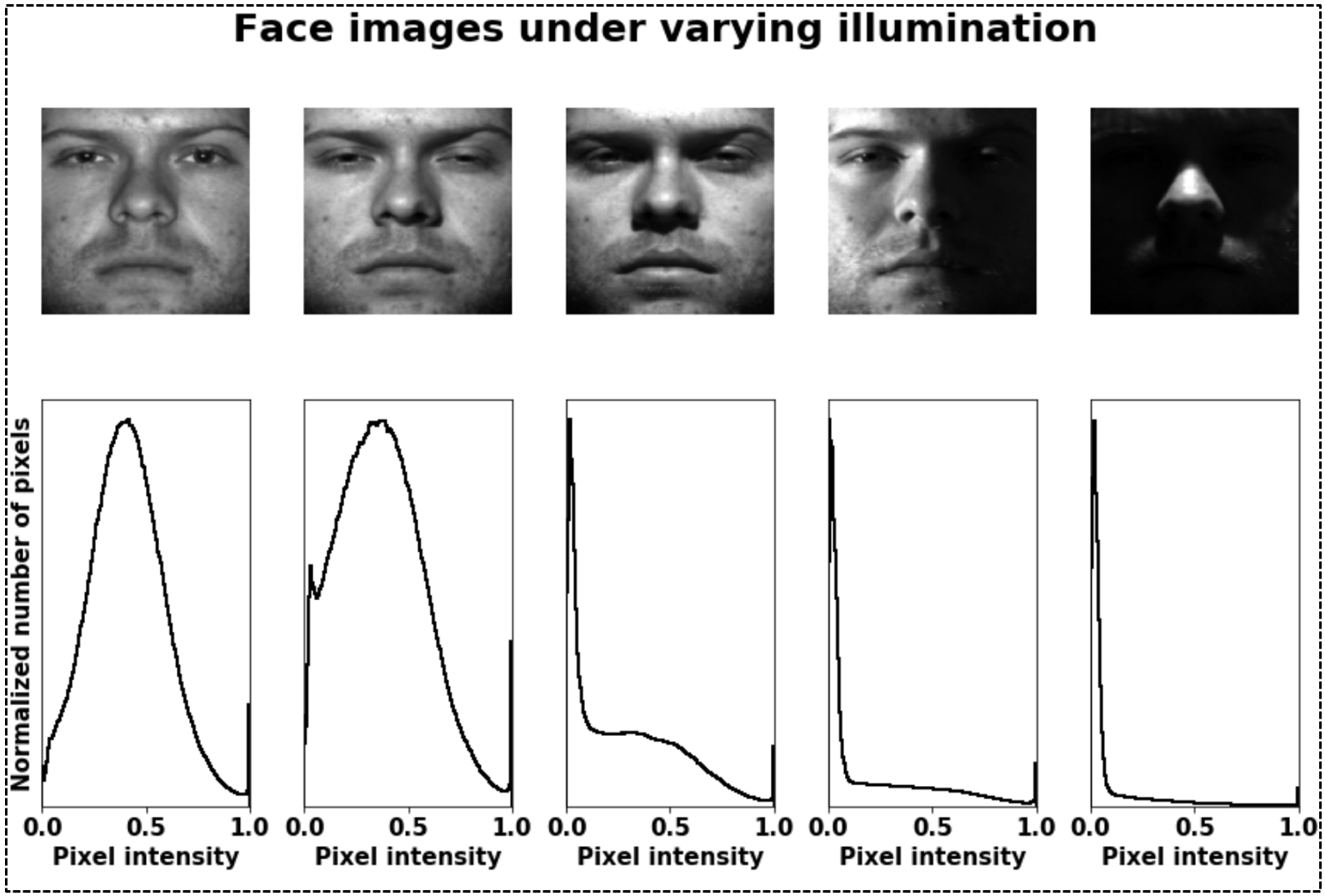}
    \caption{Top row (from left to right) demonstrates images of a subject from 5 different image subsets, which are under varying illumination conditions in the Extended Yale Face dataset~\cite{georghiades2001few}. The corresponding histogram of each image subset is shown in the bottom row.}
    \label{fig:images_examples}
\end{figure}

Automated face recognition is a necessary task for many machine-human interaction applications. Illumination variations can cause significant appearance changes for the same person and significantly affect recognition accuracy. Several pioneer studies observed that variations among images of the same person owing to variable lighting are can appear to be larger than those owing to change in identity~\cite{belhumeur1997eigenfaces, adini1997face}. One example is illustrated by the well-known Yale B Extended Face database~\cite{georghiades2001few}, as shown in Fig.~\ref{fig:images_examples}. The top row of Fig.~\ref{fig:images_examples} shows images of the same face acquired with different lighting conditions. The bottom row shows  the corresponding histograms of the corresponding pixel intensities, which change dramatically due to varying illumination conditions. As the figure shows, identifying a person when illumination changes are drastic can be challenging.

Researchers have investigated many approaches to address illumination issues for face recognition~\cite{lai2015multiscale, chen2004illumination, zhang2009face, wang2011illumination, qawaqneh2017deep, he2016deep, simonyan2014very, huang2017densely, dalal2005histograms, zhu2013logarithm, georghiades2001few, basri2003lambertian, ho2005effect, wright2008robust}. These approaches can be broadly categorized into three sub types: invariant feature extraction, 3D face modeling, and data augmentation. Some methods can include elements from more than one category. Illumination-invariant feature extraction methods aim to eliminate lighting effects by holistic decomposition~\cite{chen2004illumination}, quotient models~\cite{zhang2009face, wang2011illumination}, or logarithm difference~\cite{lai2015multiscale}, for example. These approaches apply image processing techniques to "normalize" images in such a way that they have robust appearances under varying illumination. Advantages of these methods include ease of implementation and computational efficiency. Often times, however, the formulation of these local image feature sets are empirical and are often lacking in theoretical understanding with respect to illumination effects. Another type of method uses multiple face images acquired under different illumination conditions to build a generative 3D face model that can approximate and render lighting variations of the face~\cite{georghiades2001few, basri2003lambertian, ho2005effect, wright2008robust}. However, these methods require many training images with different lighting conditions to construct such a generative model. More recently, deep learning methods have been employed for face recognition~\cite{qawaqneh2017deep, he2016deep, simonyan2014very, huang2017densely}. These end-to-end solutions leverage a large number of training images and can result in increased performance in standard face recognition tasks~\cite{parkhi2015deep}. However they typically require a large amount of data to be effective. When only a limited number of training samples are available, one may use data augmentation approach (e.g., brightness and contrast variations~\cite{shorten2019survey}) to expand the training samples for deep learning-based approaches. However, as demonstrated in the experiments below, the performance improvements are still somewhat limited, especially when illumination effects are severe.

Local image gradient-based feature sets have been used for pattern recognition including face recognition. Numerous methods following this approach have been developed such as SIFT~\cite{lowe1999object}, HoG~\cite{felzenszwalb2009object}, and LBP~\cite{ahonen2006face}. The idea behind these is to first divide images into small local patches and then compute  corresponding representations regarding local gradient information within each local image patch. These low-level local gradient-based descriptive feature sets are favored for face analysis, in part because the spatial differentiation operation naturally eliminates additive constants (i.e. brighness changes). Furthermore, splitting a face image into patches allows classification methods to be more robust with respect to slight misalignment~\cite{lowe1999object, felzenszwalb2009object}. In addition, the assumption that illumination variations in neighboring regions is smooth is widely used for lighting-invariant face analysis~\cite{chen2009wld, lai2015multiscale, chen2004illumination, zhang2009face, wang2011illumination}. Thus local low-level features are capable of providing robustness to illumination variations~\cite{sariyanidi2014automatic}. 

Our work presented here follows the same line of reasoning. We leverage local patch-wise image gradient measurements to form a new transport-based image descriptive feature set. To be precise, our proposed method divides an image into multiple local image patches, computes the corresponding local image gradients for each image patch, and explicitly represents the local image gradients as 2D discrete distributions. We illustrate that varying illumination conditions lead to certain types of deformations of local 2D discrete distributions within an image patch. Taking this knowledge into account, we construct a novel local low-level image descriptor using a sliced-Wasserstein metric computed with the aid of the Radon Cumulative Distribution Transform (R-CDT) \cite{ kolouri2015radon, shifat2020radon}. Using certain `convexifying' properties of the R-CDT transform, we are able to build a classifier that is invariant to certain deformations of the gradient distribution caused by changing illumination conditions. We show mathematically that the local (patch wise) image gradient distribution observed under certain illumination variations forms a convex set in R-CDT space and can thus be linearly separated. Equipped with these mathematical properties, we construct a subspace learning-based classifier to perform illumination-invariant face recognition. Experiment evaluations demonstrate that the proposed method achieves competitive performance among comparable approaches in three different face illumination datasets.

\begin{figure*}[ht]
\centering
\includegraphics[width=0.99\textwidth]{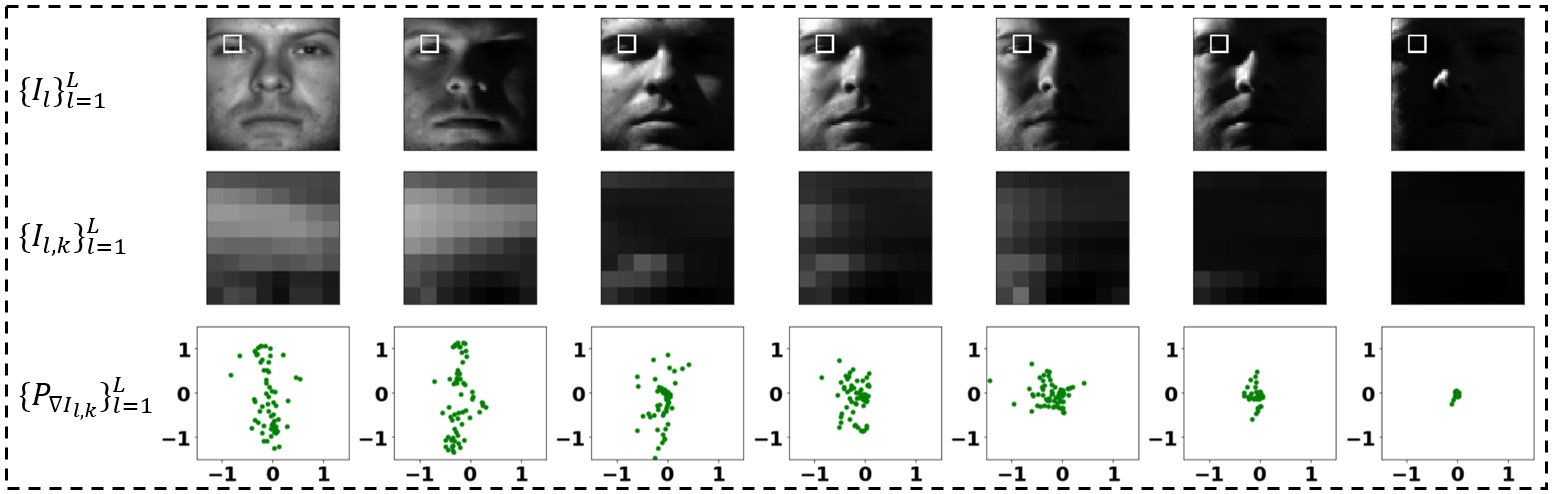}
\caption{Top row shows face images $\{I_l\}^L_{l=1}$ under various real-world illumination conditions. Middle row shows $k_\text{th}$ patches of face images $\{I_{l,k}\}^L_{l=1}$. Bottom row shows that local 2d discrete gradient distributions $\{P_{\nabla I_{l,k}}\}^L_{l=1}$ experience a combination of several types of deformations such as translation, scaling, skewing, rotation, and nonrigid deformations when illumination intensity variation present.}
\label{fig:lighting-real-world-scaling-examples}
\end{figure*}

\section{Related work} 

As mentioned above, face classification methods robust to illumination changes can be classified into three main types of approaches: (1) illumination invariant feature extraction, (2) 3D face modeling, and (3) deep learning approaches with data augmentation. Most lie on the spectrum of illumination invariant feature extraction, which refers to approaches to "remove" illumination effects using illumination normalization~\cite{gonzalez2002digital}, holistic decomposition~\cite{chen2004illumination}, quotient models~\cite{zhang2009face, wang2011illumination}, and logarithm difference~\cite{lai2015multiscale}. One straight forward approach is to employ the log transform for illumination normalization~\cite{gonzalez2002digital}, followed by lighting invariant feature extraction~\cite{chen2004illumination,zhu2013logarithm}. More specifically, Zhu $et.~al.$ first apply the log transform and then compute HoG features for face recognition~\cite{zhu2013logarithm}. Similarly, Chen $et.~al.$ use the log transform and then decompose the image into high-frequency and low-frequency components keeping only high-frequency components for face recognition~\cite{chen2004illumination}. Lai $et.~al.$ calculate the difference between neighboring pixel values in log transformation domain, formulating the logarithm-difference edge map. Then based on the size of the neighborhood, different scales of edge maps are computed and aggregated to represent each face~\cite{lai2015multiscale}. Other approaches such as WebberFace~\cite{wang2011illumination} and GradientFace~\cite{zhang2009face} are quotient-based models. To be precise, quotient models seek to represent images in such way that the current pixel value in the new representation is the ratio between the difference of current pixel and its neighboring pixel to the current pixel value in original image. However, although these methods achieve  excellent results on some datasets, they are often not effective for illumination variations that include shadows. It is worth noting that most of these approaches focus on using gradient information by directly computing differences between local neighboring pixels~\cite{zhang2009face,zhu2013logarithm}, demonstrating that the effectiveness of gradient information for illumination invariant face representation. In addition, as multiple methods calculate representations in a neighbouring region (image patch), they also implicitly rely on the assumption that the illumination variations are locally smooth.

Another research direction is to use a set of images acquired under varying lighting conditions to build a 3D face model that can render all possible illumination variations. Several researchers studied the properties of learned subspace models such as convexity and dimension \cite{epstein19955,georghiades2001few,ramamoorthi2002analytic}. However, these approaches require many training images acquired under different lighting variations to learn the required low-dimensional subspaces. Finally, deep learning-based methods such as VGGface~\cite{qawaqneh2017deep} and others~\cite{parkhi2015deep, he2016deep, simonyan2014very, huang2017densely} have become increasingly popular. These typically require a large amount of data for training. To overcome these limitations, many  data augmentation strategies to increase number of training samples can be used~\cite{shorten2019survey}. However, when a limited amount of training data is available, even with data augmentation techniques~\cite{shorten2019survey}, deep learning-based methods tend to have relatively poor generalization properties~\cite{azulay2018deep}.

\begin{table}[ht]
\centering
\resizebox{0.5\textwidth}{!}{%
\begin{tabular}{clclllllll}
\multicolumn{10}{c}{Notations}                 \\ \hline
\multicolumn{2}{c}{$\Omega$, $\Omega_k$} & \multicolumn{8}{c}{image coordinate domain, coordinate domain for patch $k$}   \\
\multicolumn{2}{c}{ $\Omega_k^N$} & \multicolumn{8}{c}{a set $\{\bx_1^k,...,\bx^k_N\}$ of $N$ pixel locations on $\Omega_k$ }   \\
\multicolumn{2}{c}{$I_k$} & \multicolumn{8}{c}{$I|_{\Omega_k}$:$k_\text{th}$ patch of image $I$}   \\

\multicolumn{2}{c}{C, L, K} & \multicolumn{8}{c}{total number of subjects, image samples, and patches}   \\
\multicolumn{2}{c}{$\delta_{z}$} & \multicolumn{8}{c}{{Dirac measure centered at $z$}}\\
\multicolumn{2}{c}{$P_{\nabla I_{k}}$} & \multicolumn{8}{c}{2D discrete gradient distribution of $I_k$}   \\
\multicolumn{2}{c}{$P_{\nabla _{\theta}I_k}$} & \multicolumn{8}{c}{ 1D discrete distribution of directional derivative of $I_k$ alone angle $\theta$ }\\
\multicolumn{2}{c}{$\mathbb{H}$} & \multicolumn{8}{c}{a subset of bijection deformations from $\R^2$ to $\R^2$}\\

\multicolumn{2}{c}{$\mathbb{H}_0$} & \multicolumn{8}{c}{set of all possible compositions of translation and scaling diffeomorphisms} \\

\multicolumn{2}{c}{$\wtheta$} & \multicolumn{8}{c}{{directional vector $[\cos \theta,\sin\theta]^T$}}\\

\multicolumn{2}{c}{$P^h_{\nabla I_k}$} & \multicolumn{8}{c}{2D discrete gradient distribution deformed (push-forwarded) from $P_{\nabla I_{k}}$ via $h\in \mH$}\\

\multicolumn{2}{c}{$\F$} & \multicolumn{8}{c}{$\transname$ for 1D discrete distributions}\\
\multicolumn{2}{c}{$\F^*$} & \multicolumn{8}{c}{$\hogname$ for 2D discrete distributions}\\
\multicolumn{2}{c}{$\widehat P_{\nabla I_{k}}$} & \multicolumn{8}{c}{$\widehat P_{\nabla I_{k}}:= \F^*\big(P_{\nabla I_{k}}\big)$}\\
\multicolumn{2}{c}{$\mathbb S^c$} & \multicolumn{8}{c}{a generative model for images of subject $c$ under illumination variations}\\
\multicolumn{2}{c}{$\V_k^c$} & \multicolumn{8}{c}{the subspace in the Discrete R-CDT domain corresponding to $k_\text{th}$ patch of subject $c$  }\\
\multicolumn{2}{c}{$d(\cdot, \cdot)$} & \multicolumn{8}{c}{ the discrete sliced Wasserstein distance between two 2D distributions}\\
\end{tabular}%
}
\end{table}
\section{Proposed solution}

We propose a novel approach for illumination invariant face recognition. The method splits the image into into multiple image patches and uses the local 2D discrete gradient distribution to represent the information in each patch. We demonstrate that illumination variations cause deformations of the local 2D discrete gradient distributions that can be modeled mathematically as well as learned from available data. Based on this observation, we propose a novel local image descriptor based on the Radon Cumulative Distribution Transform (R-CDT) \cite{shifat2020radon} of the local gradient distribution. We model the set of all local image patches observed under certain illumination variations as a subspace in $\hogname$ domain. Given a test patch, the $\hogname$ of the test patch is computed, and the distance to the corresponding patch subspace for each subject in the database is computed. The average nearest subspace subject is then chosen as the classification result.

\subsection{Effects of varying illumination conditions on local 2D gradient distributions}
\label{sec:problemIntro}

\subsubsection{Notation}

\begin{figure}[H]
\centering
  \includegraphics[width=0.48\textwidth]{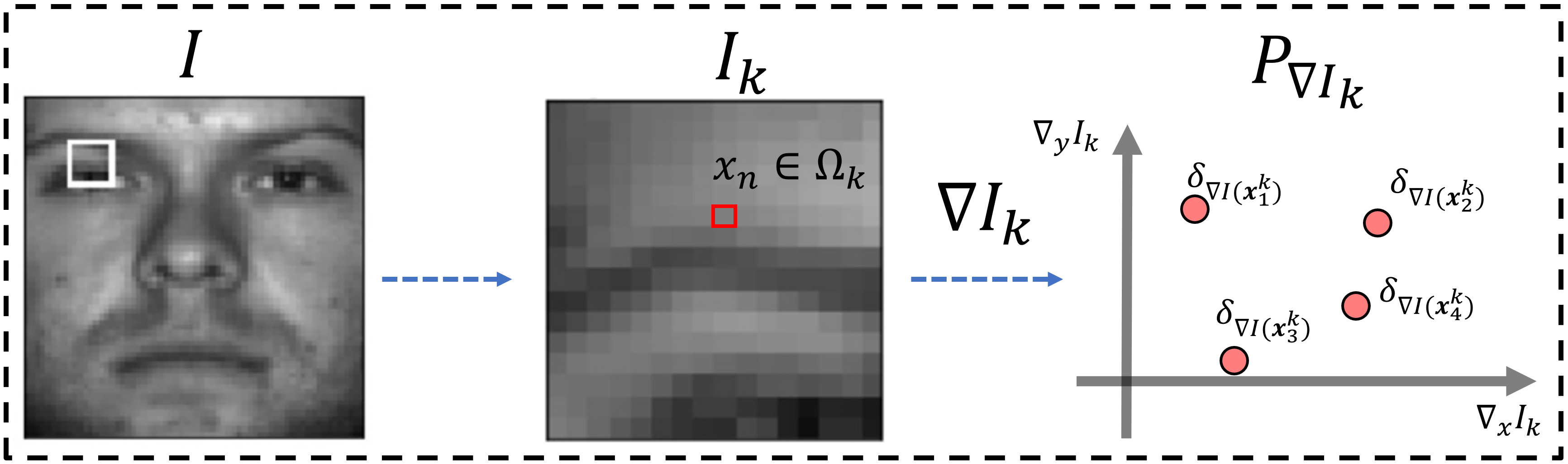}
\caption{From image space to local 2d discrete gradient distribution: given an image patch $\nImagePatch$, one can compute its 2d discrete gradient distribution $\nDistriP$. Please note that here an illustration example is plotted to show $\nDistriP$.} 
\label{fig:image2GradientDis}
\end{figure}

An image $I: \Omega\rightarrow \R_{+}$ can be thought as a mapping from the unit square $\Omega = [0,1]\times [0,1]$ to the set  $\R_{+}$ of non-negative real numbers. Let $\Omega_k \subset \Omega$ refer to a set of pixel coordinates in the $k^{\text{th}}$ neighborhood in $\Omega$ and  $\Omega_k^N=\{\bx_1^k,...,\bx_N^k\}\subset \Omega$ is the set of $N$ pixel locations for patch $k$.  

Given an image patch $\nImagePatch: \Omega_k \rightarrow \R_{+}$,  the corresponding 2D discrete gradient distribution $\nDistriP$ for the $k_\text{th}$ patch is defined as 

\begin{equation}
\nDistriP := \frac{1}{N}\sum_{\bx\in \Omega_k^N}\delta_{\nabla I(\bx)}.
\label{eq-DiscreteDistributon}
\end{equation}

\noindent An example image $I$, image patch $I_k$, and corresponding distribution $P_{\nabla I_k}$ are shown in Figure~\ref{fig:image2GradientDis}. Note our use of measure theoretic notation $\delta_z$ for a delta mass positioned at $z$ (see Appendix \ref{append:dirac} for a brief introduction to Dirac measures.)\\

\subsubsection{Effects of illumination changes on 2D gradient distributions}

An example of one image patch observed under different illumination conditions is shown in Fig.~\ref{fig:lighting-real-world-scaling-examples}. The example shows that the gradient distribution $P_{\nabla I_k}$ undergoes deformations that may include translation, scaling, skewing, rotation, and nonrigid deformations according to the reflectance properties of the object being imaged, its three dimensional configuration, and specific illumination conditions. Here we propose a transport-based way of modeling such deformations.

Given the local gradient distribution $\nDistriP$ of a subject observed under a certain illumination condition,  the corresponding patch gradient distribution of the same subject after some illumination changes can be modeled by a bijective transformation $h$ as:
\begin{equation}
	\nDistriPh := {h}_{\sharp} {\nDistriP}=\frac{1}{N}\sum_{\bx\in \Omega_k^N}\delta_{h(\nabla I(\bx))}, \quad h\in \mH
	\label{eq:transport_model}
\end{equation}
where $\mathbb H$ is a set of  bijective deformations from $\R^2$ to $\R^2$ that are particular to certain illumination assumptions (see more below). We denote the  the push forward of distribution $\nDistriP$ as ${h}_{\sharp} {\nDistriP}$ (see Appendix \ref{1d-cdt-composition-property} for definitions and properties of the push-forward operation). Moreover,  we model the set of gradient distributions of a subject image patch $I_k$ under various illumination variations with the following template-deformation based generative model:  
\begin{equation}
    \nSetH = \{\nDistriPh \mid h\in \mathbb{H}\}.
	\label{eq:generativeModelTemplate}
\end{equation}
A particular model for set $\mathbb{H}$ that describes illumination effects is proposed below. Here $\nSetH$ refers to the set of all possible observations of $\nDistriPh$. We will utilize geometric properties of $\nSetH$ in our problem statement and solution described below. \\

\subsubsection{Modeling $\mathbb{H}$: local illumination changes and patch generative model}
\label{sec:affineModel}

The set $\mH$ of deformations corresponding to the set of illumination variations within a subject class can be hard to specify in general. However, we can make some reasonable assumptions and propose an approximate model that can enable robust classification. Let the size (area) of neighborhood $\Omega_k$ be denoted as $|\Omega_k|$. We then postulate that, for small enough neighborhood size $|\Omega_k|$, we have that
\begin{equation}
    h(\mathbf{z}) \sim \alpha \mathbf{z} + \mathbf{b} 
    \label{eq:h_model}
\end{equation}
where $\mathbf{z}$ is a gradient coordinate (i.e. $\mathbf{z} = \nabla I_k(x), x \in \Omega_k$), and where $\alpha \in \mathbb{R}$ is an unknown scaling function, and $\mathbf{b} \in \mathbb{R}^2$ is an unknown translation vector. The illumination model above can be derived from the assumption that within a local patch, illumination variations within that patch can be expressed as 
\begin{equation}
  \alpha  I_{k}(\bx) + \beta + \bb^T\bx, \mathbf{x} \in \Omega_k.
\label{eq:imageSpace}
\end{equation}
under the assumption that $|\Omega_k|$ is small. In the equation above, $\alpha>0$ is known as contrast, $\beta$ illumination intensity, and $\mathbf{b}$ is a linear gradient (caused by illumination at an angle, or potentially shadows) superimposed on the image. Under the assumption of small neighborhood $|\Omega_k|$, the model we propose in equation \eqref{eq:h_model} can be understood as the gradient of \eqref{eq:imageSpace}, and thus can be understood in terms of illumination intensity, contrast, and linear illumination gradient. We note that similar smoothness assumptions are widely used for illumination-invariant face recognition \cite{chen2009wld, lai2015multiscale, chen2004illumination, zhang2009face, wang2011illumination}. 

Based on \eqref{eq:h_model} above we propose a specific model for the set of bijections $\mathbb{H}$ that cause gradient deformations (pushforward) of a given gradient distribution:
\begin{equation}
\mathbb H_0 = \{h(\bz) = \alpha\bz +\bb \mid a>0, \bb\in \R^2\}.
	\label{eq:affineModel}
\end{equation}
Finally, using the notation established earlier, we express the set of gradient distributions observed under an unknown illumination function $h \in \mathbb{H}_0$
as 
\begin{equation}
	\nDistriPh :=\frac{1}{N}\sum_{\bx\in \nOmegaNPoints}\delta_{h(\nabla I(\bx))} = \frac{1}{N}\sum_{\bx\in \nOmegaNPoints}\delta_{\alpha \nabla I(\bx) + \textbf{b}}.
	\label{eq:translateScaleModel}
\end{equation}
and the set of all possible observations as
\begin{equation}
    \nSetHNot= \{\nDistriPh = {h}_{\sharp} {\nDistriP} \mid h\in \mathbb{H}_0\}
	\label{eq:generativeModelTemplateAffine}.
\end{equation}
We denote the set defined in equation \eqref{eq:generativeModelTemplateAffine} as a model for the gradient distribution of patch $k$ under illumination model $\mathbb{H}_0$. Specifically $\mathbb{H}_0$ is capable of isotropically scaling and translating a given distribution $P_{\nabla I_k}$. Figures \ref{fig:scaleGradients} and \ref{fig:linearGradients} in Section.\ref{sec:simulatedAffine} (appendix) provide visualizations of these variations. Leveraging the knowledge present in equation~\eqref{eq:imageSpace} and equation~\eqref{eq:generativeModelTemplateAffine}, we propose the following patch-wise affine generative model and classification problem using gradient distributions for face images under varying illumination conditions.\\

\noindent\textbf{Problem statement}: Let the local illumination-based generative model for images pertaining to subject (class) $c=1,\cdots,C$ be defined as:
\begin{equation}
\label{eq:GenModelc}
   \mathbb S^c = \Big\{ I^{c,j} \  \begin{tabular}{|l}
      $I^{c,j}_k(\bx) = \alpha_k^j I^c_k(\bx) + \beta_k^j + \bb_k^j\cdot \bx, $ \\
      $\bx \in \Omega_k,\alpha_k^j>0, \beta_k^j \in \R, \bb_k^j\in \R^2$,\\
      $k=1,...,K$
\end{tabular} \Big\},
\end{equation}
 where $I^c$ refers to an (unknown) template image for subject (class) $c$. This model defines an infinite set, whose elements $I^{c,j}$ can be generated by applying the illumination model \eqref{eq:imageSpace} on each patch $k$ of $I^c$ independently. In other words, the generative model is flexible to allow each patch $k$ to contain its own contrast, brightness, and gradient vector ($\alpha_k^j,\beta_k^j,\bb_k^j$) parameters. These illumination parameters are unknown for any given photograph. Given $L$ training images $\{I^{c,l}\}_{l=1}^L \subseteq \mathbb S^c$ from each class $c=1,...,C$ where $\mathbb S^{c} \cap \mathbb S^{c^{\prime}} =\emptyset$ for all $c\neq c^{\prime}$, determine the class of an unknown image $I^t$ obtained from the same generative model.
 
 It is not hard to see that  $\nabla I^{c,j}_k = h_k^j \circ \nabla I^c_k$ where $h_k^j(\bz)=\alpha_k^j \bz+\bb_k^j$. In other words, the patch gradient distributions for each class $c$ satisfy the affine generative model stated in \eqref{eq:generativeModelTemplateAffine}, i.e., 
  \begin{equation}\label{eq:GradGenModelc}
  \{P_{\nabla I_k^{c,j}}\mid I^{c,j}\in \mathbb S^c\}= 	\mP_{\mH_0,k}^c:=\{P_{\nabla I_k^c}^h \mid h\in \mH_0\}.
  \end{equation}

\subsection{Proposed solution}

We propose a straightforward, non iterative, solution to the classification problem stated above. The solution is inspired on prior work on classification of distributions \cite{park2018cumulative,kolouri2015radon,shifat2020radon} and utilizes the fact that the gradient distribution $P_{\nabla I^t_k}$ for patch $k$ is an element of $\mathbb{P}^c_{\mathbb{H}_0,k}$, with $c$ unknown. In other words $P_{\nabla I^t_k} \in \mathbb{P}^c_{\mathbb{H}_0,k}$, with $\mathbb{P}^c_{\mathbb{H}_0,k} = \{{h}_{\sharp} P_{\nabla I^c_k} \mid h\in \mathbb{H}_0\}$, following the definition in equation \eqref{eq:generativeModelTemplateAffine}, for some unknown $c$. We use a distance function $d(P_{\nabla I^t_k},\mathbb{P}^c_{\mathbb{H}_0,k})$ that measures the Sliced Wasserstein Distance \cite{kolouri2017optimal,rabin2011wasserstein} between $P_{\nabla I^t_k}$ and the nearest point in set $\mathbb{P}^c_{\mathbb{H}_0,k}$ to compute the solution of the classification problem stated above as:
\begin{equation}\label{eq: argminProb}
    c^* = \arg\min_c \sum_{k=1}^{K} d^2(P_{\nabla I^t_k},\mathbb{P}^c_{\mathbb{H}_0,k}).
\end{equation}
It is easy to show that the minimization above obtains the correct solution to the problem statement above, provided that for at least least one $k$ we have that $\mathbb{P}^{c}_{\mathbb{H}_0,k} \cap \mathbb{P}^{c^{\prime}}_{\mathbb{H}_0,k} =\emptyset$ whenever $c\neq c^{\prime}$. Below we show how we can estimate $d^2(P_{\nabla I^t_k},\mathbb{P}^c_{\mathbb{H}_0,k})$ with the aid of a newly introduced operation which we will call the Discrete Radon Cumulative Distribution transform (Discrete R-CDT). \\

\subsubsection{Sliced-Wasserstein representation of local 2D discrete distribution}

\begin{figure}
\centering
  \includegraphics[width=0.49\textwidth]{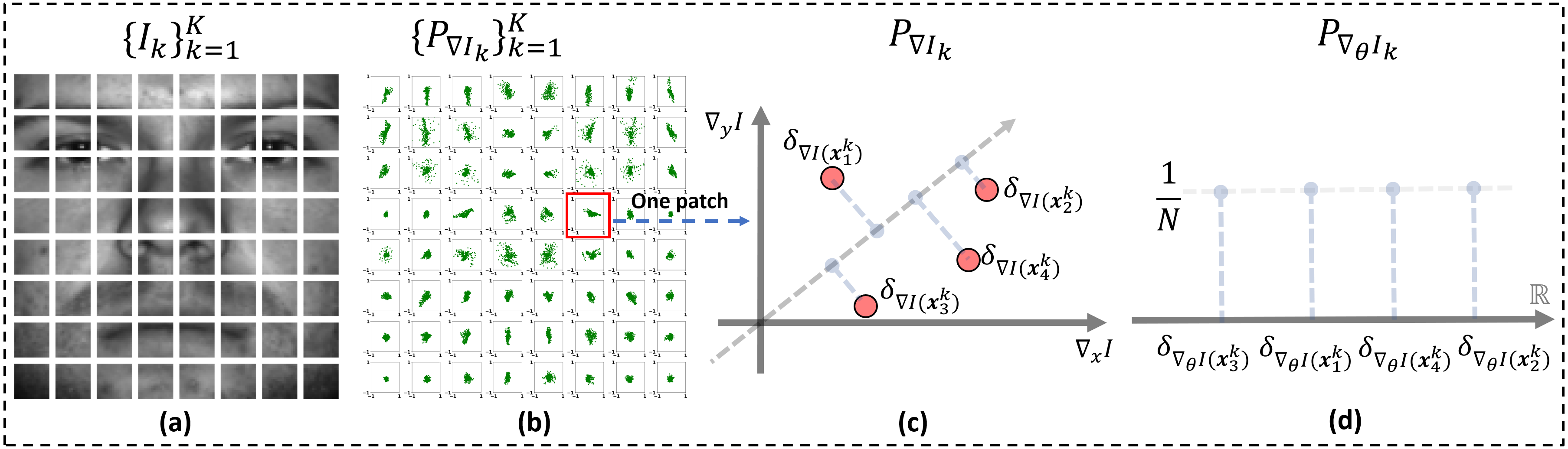}
\caption{An illustration example of computing a sliced one-dimensional discrete distribution for one image patch: (a) a face image consisted of $K$ image patches $\{I_k\}_{k=1}^K$; (b) 2D discrete gradient distributions $\{P_{\nabla I_k}\}_{k=1}^K$ corresponding to image patches $\{I_k\}_{k=1}^K$; (c) a 2D discrete gradient distribution $P_{\nabla I_k}$ for a particular image patch; (d) sliced one-dimensional representation of $P_{\nabla I_k}$. Please note that here subplots (c) and (d) are only used for illustration purpose.}
\label{fig:sliced-1d}
\end{figure}

Given an image $I$ where the patch-wise representation is $\{I_{k}\}^K_{k=1}$, we compute its local gradient distribution $P_{\nabla I_k}$ as defined in equation \eqref{eq-DiscreteDistributon} and use a modified version of the Radon Cumulative Distribution Transform (R-CDT) to represent $P_{\nabla I_k}$. The idea is to use a set of one-dimensional discrete distributions by ``slicing" $P_{\nabla I_k}$ along different angles. 

To begin, we note that one can compute the 1D distribution $P_{\nabla_{\theta} I^c_k}$ of projected gradients via 
\begin{equation}\label{eq:proj_distr}
	P_{\nabla_{\theta} I_k}=\frac{1}{N}\sum\limits_{\bx\in\Omega_k^N} \delta_{\nabla I(\bx)\cdot \wtheta},
\end{equation}
where $\wtheta=(\cos\theta,\sin\theta)^T$ is a unit vector in the direction of $\theta$. Inspired on earlier work on the CDT \cite{park2018cumulative} and R-CDT \cite{kolouri2015radon}, we then define the $\transname$ for one-dimensional discrete probability distributions:\\

\noindent\textit{Definition 3.1} (Discrete Cumulative Distribution Transform): 
Let $\mP_N(\R):= \{P_Z = \frac{1}{N}\sum_{i=1}^N\delta_{z_i}\mid Z =\{ z_i\}_{i=1}^N\subset \R\}$ be the set of discrete probability distributions concentrated on $N$ points on $\R$.
The {\hogname}  $\F: \mP_N(\R) \rightarrow  \R^N$ is defined 
\begin{equation}\label{eq:dcdtdef}
	\F(P_Z) = \mathcal P\big[z_1,..., z_N\big]^T = \big[\widetilde z_1,..., \widetilde z_N\big]^T,
\end{equation}
where $\mathcal{P}$ is a permutation matrix such that $\widetilde z_1\leq \cdots\leq \widetilde z_N$.\\

One can show that $\F$ is an isometric embedding from the 1D discrete probability distribution space with the Wasserstein metric to the transform space with the Euclidean distance (see e.g., ~\cite{ kolouri2017optimal, santambrogio2015optimal}). That is, given two 1D discrete distributions $P_{Z^{(1)}}$ and $P_{Z^{(2)}}$ in $\mP(\R)$, we have that the Wasserstein distance between them is given by
\begin{equation}\label{eq:dcdtembedding}
	W_2(P_{Z^{(1)}}, P_{Z^{(2)}}) = \sqrt{\frac{1}{N}||\F(P_{Z^{(1)}})-\F(P_{Z^{(2)}})||^2}
\end{equation}
where $||\cdot||$ denotes the Euclidean distance on $\R^N$. In particular, for a directional derivative distribution $P_{\nabla _{\theta}I_{k}}\in \mP_N(\R)$, we have that
\begin{equation}
    \F(P_{\nabla _{\theta}I_{k}}):= \mathcal{P}_{\theta} \begin{bmatrix}
                \nabla_\theta I(\bx_1^k) \\ \nabla_\theta I(\bx_2^k) \\ \vdots \\ \nabla_\theta I(\bx_N^k)
            \end{bmatrix}
             = \begin{bmatrix}
                \widetilde{\nabla_\theta I(\bx_1^k)} \\ \widetilde{\nabla_\theta I(\bx_2^k)} \\ \vdots \\ \widetilde{\nabla_\theta I(\bx_N^k)}
            \end{bmatrix},
\label{eq.1dcdt}
\end{equation}
where $\nabla_\theta I(\bx_i^k) = \nabla I(\bx_i^k)\cdot \wtheta$ and 
 $\mathcal{P}_{\theta}$ is a permutation matrix such that $\widetilde{\nabla_\theta I({{\bx}_1}^k)} \leq \widetilde{\nabla_\theta I({{\bx}_2}^k)} \leq \cdots \leq \widetilde{\nabla_\theta I({{\bx}_N}^k)}$. 
 In other words, the $\transname$ $\F$ takes the one-dimensional discrete distribution $P_{\nabla_{\theta} I_{k}}$ as input and outputs a vector which are concentration locations of the discrete distribution in an increasing order.

Combining the "slicing" idea shown in \eqref{eq:proj_distr} and the $\transname$ defined in \eqref{eq:dcdtdef}, 
we define the $\hogname$ for 2D discrete distributions:\\ 

\noindent\textit{Definition 3.2}{ (\hogname)}:
Let $\mP_N(\R^2):= \{P_{\bZ} = \frac{1}{N}\sum_{i=1}^N\delta_{\mathbf z_i}\mid \bZ =\{\mathbf z_i\}_{i=1}^N\subset \R^2\}$ be the set of discrete probability distributions concentrated on $N$ points on $\R^2$.
The {\hogname}  $\HOG: \mP_N(\R^2) \rightarrow  \transpace:= \{v: [0,\pi) \rightarrow \R^N\}$, denoted as $ \widehat P_{\bZ}:=\HOG (P_{\bZ}) $, is defined such that for each $\theta \in [0,\pi)$: 
\begin{equation}
  \Big( \HOG (P_{\bZ})\Big)(\theta) = \F\Big(P_{\bZ_{\theta}}\Big),
\end{equation}
where $P_{\bZ_{\theta}}= \frac{1}{N}\sum_{i=1}^N\delta_{\mathbf z_i\cdot \wtheta}$ is a one-dimensional distribution concentrated on the projected values of $Z$ onto the directional vector $\wtheta$.\\

With \eqref{eq:dcdtembedding} in mind, we define the Discrete Sliced Wasserstein Distance~\cite{rabin2011wasserstein}
\begin{align}\label{eq: dsw}
	d\big(P_{\bZ^{(1)}},P_{\bZ^{(2)}}\big):&=||\widehat P_{\bZ^{(1)}}-\widehat P_{\bZ^{(2)}}||_{\ltwog} \notag \\
	&=||\F^{*}\big(P_{\bZ^{(1)}}\big)-\F^{*}\big(P_{\bZ^{(2)}}\big)||_{\ltwog} \notag\\
	&= \sqrt{\int_0^{\pi}||\F\big(P_{\bZ^{(1)}_{\theta}}\big)- \F\big(P_{\bZ^{(2)}_{\theta}}\big)||^2 d\theta}
\end{align}
where $||\cdot||$ denotes the Euclidean norm in $\R^N$. In particular, when $P_{\bZ}$ is a gradient distribution, say $P_{\bZ} = P_{\nabla I_{k}}$, then
\begin{equation}
   \widehat P_{\nabla I_{k}}(\theta)= \Big( \HOG (P_{\nabla I_{k}})\Big)(\theta) = \F(P_{{\grad_{\theta} I_{k}}}).
\label{2dRCDTHoGDef}
\end{equation}
In other words, applying the Discrete R-CDT (operator $\F^*$) on the local 2D discrete distribution (i.e, $P_{\nabla I^c_{1,k}}$) equates to applying \transname (i.e., $\F$) on a collection of projected one-dimensional representations\footnote{This projection is similar to the sliced projections in the Radon transform, and this is why we have "Radon" (R) in the name of this new transform.} of (e.g., $P_{\nabla_{\theta} I_{k}}$), indexed by  $\theta\in [0, \pi)$. The corresponding Discrete Sliced Wasserstein distance between two patch gradient distributions  $P_{\nabla I_k^t}$ and $P_{\nabla I_k^{c,j}}$ is 
\begin{align}
d(P_{\nabla I_k^t}, P_{\nabla I_k^{c,j}}) &= ||\widehat P_{\nabla I^t_k}-\widehat P_{\nabla I^{c,j}_k}||_{\ltwog} \notag\\
&= \sqrt{\int_0^{\pi}||\F\big(P_{\nabla_{\theta} I^t_k}\big)- \F\big(P_{\nabla_{\theta} I^{c,j}_k}\big)||^2 d\theta}.
\end{align}
The minimization problem \eqref{eq: argminProb} is hence equivalent to 
\begin{align}\label{eq: argminSWmetric}
    c^* &= \arg\min_c \sum_{k=1}^{K} d(P_{\nabla I^t_k},\mathbb{P}^c_{\mathbb{H}_0,k}) \nonumber\\
    &= \arg\min_c \sum_{k=1}^{K}\min_{h_k\in \mH_0} ||\widehat P_{\nabla I^t_k}-\widehat P^{h_k}_{\nabla I^{c}_k}||_{\ltwog}.
\end{align}

\noindent{Remark 3.1:}
 In practice, we take $\theta$ from a finite set $\{\theta_1, ...,\theta_m\}$ for some positive integer $m$,  $\widehat P_{\nabla I_{k}}=\HOG (P_{\nabla I_{k}})$ can be represented by a matrix of size $N\times m$ and reshaped as a long vector of length $m*N$. \\

In summary, the {\hogname} takes 2D discrete distribution as input and outputs a sequence of vectors indexed by $\theta$ in some finite set.\\

\subsubsection{Nearest subspace learning}
\label{sec-ns}
Next we leverage the generative model as in \eqref{eq:GenModelc} together with the \hogname~ $\F^{*}$ to form a nearest  subspace classification method to facilitate the classification strategy in \eqref{eq: argminProb}.  It is not hard to see that $\F^{*}\big(\mP_{\mH_0,k}^c\big)$ is convex, meaning that $\lambda \F^{*}\big(P_{\nabla I_k^c}^{h_1}\big) + (1-\lambda) \F^{*}\big(P_{\nabla I_k^c}^{h_2}\big)$ lies in $\F^{*}\big(\nSetHNot\big)$ for all $\lambda\in [0,1]$ and $h_1,h_2\in \mH_0$ (see Section.~\ref{sec:2DConvexityProof} in the appendix for a proof). Indeed, the deformations $\mathbb{H}_0$ in 2D discrete distribution space also cause the corresponding translation and scaling effects in $\hogname$  space: 
\begin{align}
    \F^*\big(P_{(\alpha\nabla I_{k} + \bb)}\big)(\theta) &= \F\big(P_{(\alpha\nabla_\theta I_{k} + \bb \cdot \wtheta)}\big) \notag\\
    &= \alpha\F(P_{\nabla_{\theta} I_{k}})+\bb \cdot \wtheta \notag\\
    &= \alpha\F^*(P_{\nabla I_{k}})(\theta)+\bb \cdot \wtheta,
\label{eq.2dTranslationScaling}
\end{align}
\noindent where the second equation follows from the composition property of the Discrete CDT transform (please see Section \ref{1d-cdt-composition-property}) and the addition on the RHS is operated entry-wise.  We then expand the convex set $
\F^{*}\big(\nSetHNot\big)$ to form a subspace $\V_k^c= \textrm{span}\big(\F^{*}(\mP_{\mH_0,k}^c)\big)$ and further assume that when  $c\neq c^{\prime}$, there exists a patch $k$ such that $	\mP_{\mH_0,k}^{c} \cap \V^{c^{\prime}}_k = \emptyset$,  which is consistent with the assumption that the image class of subject $c$  will not overlap with images of a different subject $c^{\prime}$ under all possible illumination variations. With the above considerations in mind, the constrained minimization problem in  \eqref{eq: argminProb} or \eqref{eq: argminSWmetric} can be modified to a simple subspace projection  problem \eqref{eq:subspaceProj}, which can be solved by basic linear algebra techniques in transform domain as shown in sections below: 
\begin{align}
	c^{*} 	&=\arg\min_{c}\min_{\widehat P\in \V_k^c} \sum_{k=1}^K ||\widehat P_{\nabla I_k^t} - \widehat P||_{\ltwog} \label{eq:subspaceProj}\\
	& = \arg\min_{c}  \sum_{k=1}^K d\big(\widehat P_{\nabla I_k^t}, \V_k^c\big)\nonumber\\
	& = \arg\min_{c} \sum_{k=1}^K d_k^c \nonumber
\end{align}
where  $d^c_k := d\big(\widehat P_{\grad I^t_k}, \V_k^{c}\big)$ is the $d(\cdot,\cdot)$ distance of $\widehat P_{\grad I^t_k}$ to the subspace $\V_k^{c}$, which can be computed in a convenient form as a least squares projection as shown below. In summary, the class is determined by the smallest distance $d^c= \sum_{k=1}^K d_k^c$. 
In particular, if a test image $I^t$ belongs to subject $c$, $\sum_{k=1}^K d_k^c  < \sum_{k=1}^K d_k^{c^{\prime}}$ for any $c^{\prime}\neq c$ (see \ref{appendix:metric} for more details).\label{sec:proposedApproch}

\paragraph{Training}
Given a total of $\{I^c_l\}_{l=1}^L$ $L$ training images for $c_{th}$ subject. Each image $I_l$  is partitioned into $K$ image patches $\{I^c_{l,k}\}^K_{k=1}$. We approximate the  subspace $\V^c_k$ using  $L$ training images $\{I^c_l\}_{l=1}^L$ by  
\begin{equation}
\V^c_k = \text{span}\Big(\{\widehat P_{\nabla I^c_{1,k}}, \cdots, \widehat P_{\nabla I^c_{L,k}}\} \cup \mathbb{U}_T\Big),
\label{eq.subspace}
\end{equation}
where $\mathbb{U}_T = \{\mu_1(n,\theta), \mu_2(n,\theta)\}$ with $\mu_1(n,\theta)=\cos{\theta}$, $\mu_2(n,\theta)=\sin{\theta}$ can be used to automatically model translation and scaling within a subject gradient distribution class by observing Equation \eqref{eq.2dTranslationScaling} (see also ~{\cite{shifat2020radon,Shifat2022invariance}}). It is worth noting that if the deformation strictly follows the set $\mathbb{H}_0$, only taking span of one transformed training example with $\mathbb{U}_T$ is necessary. However, in reality it is often the case that more complicated illumination effects than defined in $\mathbb{H}_0$ are present. We can enhance $\mathbb{H}_0$ by using any available training images using eq.~(\ref{eq.subspace}) to take span of multiple transformed training examples with $\mathbb{U}_T$. This technique allows the proposed method to learn from more complicated lighting variations. A more detailed discussion is provided in Section~\ref{sec:futurework}.

Recall that in our algorithm,  each $\widehat P_{\nabla I^c_{l,k}}$ in the spanning set can be discretized into a vector in $\R^{mN}$ where $m$ is number of angles used in practice (see Remark 3.1). For simplicity, we abuse the notation and use the same symbol for the discretized version of $\widehat P_{\nabla I^c_{l,k}}$ in algorithms, which will be clear from the context. Similarly one can discretize functions in $\mathbb U_T$  by evaluation on the grid $\{1,2,..., N\}\times \{\theta_1,...,\theta_m\}$ and reshape them as a vectors of length $m*N$, which we abuse notation and denote as $\mathbb U_T$\footnote{We also abuse the notation and denote the subspace of the discretized spanning functions in \eqref{eq.subspace} as $\V^c_k$.}. Now we summarize  the training algorithm in the following steps: for each class $c$ and each patch $k$,
\begin{enumerate}
\item 	Compute the transforms $\widehat P_{\nabla I^c_{1,k}}, \cdots, \widehat P_{\nabla I^c_{L,k}}$ corresponding to the training images
\item  Use Principal Component Analysis (PCA)~\cite{bishop2006pattern}, keeping enough components to retain 99\% of the training data variance, to orthogonalize $\{\widehat P_{\nabla I^c_{1,k}}, \cdots, \widehat P_{\nabla I^c_{L,k}}\} \cup U_T$ to obtain a set of orthonormal basis vectors $\{v_{1,k}^c,v_{2,k}^c,\cdots\}$ and form a matrix $B^c_k$ with $\{v_{1,k}^c, v_{2,k}^c,\cdots\}$ as its columns:
\begin{equation}
	B^c_k = [v_{1,k}^c, v_{2,k}^c,\cdots].
\end{equation}
When enough training data is available, we split the training data into training and validation sets, and choose the smallest number of components that allow for highest classification accuracy on the validation set.\\
\end{enumerate}

\paragraph{Testing} Given a testing image $I^t$, the first step is to segment $I^t$ into $K$ image patches $\{I^t_k\}^K_{k=1}$ and calculate the $\hogname$ representation $\{\widehat P_{\nabla I^t_{k}}\}^K_{k=1}$. Then, for each  image patch $k$, we compute the distance 
\begin{equation}
	d(\widehat P_{\nabla I^t_{k}},\V^c_k) = ||\widehat P_{\nabla I^t_{k}}- B^c_k(B^c_k)^T \widehat P_{\nabla I^t_{k}})||,
\end{equation}
where $(B^c_k)^T $ is the transpose of matrix $B^c_k$ and $||\cdot||$ denote the Euclidean norm.
 Finally, we compute ${d^c}$ by summing the distance contribution $d^c_k$ and search for the nearest subspace as the classification result via:
\begin{equation}
\label{eq.dis}
    \arg\min_c d^c= \sum_{k=1}^K d^c_k.
\end{equation}
 
\section{Experiments}

We evaluate the proposed method on three different face recognition datasets with illumination variations: (1) Extended Yale Face Database B that has 38 different subjects under 68 types of lighting variations~\cite{georghiades2001few}; (2) AR Face Dataset that has 100 subjects under 4 illumination conditions~\cite{martinez1998ar}; (3) CAS-PEAL dataset that has 233 subjects under more than 9 lighting conditions~\cite{gao2007cas}. We perform comparisons between the proposed method and other illumination-invariant face recognition algorithms~\cite{chen2004illumination, lai2015multiscale, zhang2009face, wang2011illumination, zhu2013logarithm,felzenszwalb2009object} as well as several deep learning based alternatives~\cite{simonyan2014very, he2016deep, huang2017densely} with illumination data augmentation strategy. Specifically, we consider three state-of-the-art deep learning models: VGGFace, ResNet-50 and DenseNet-121. We use 90\% and 10\% of the original training data for training and validation, respectively. Validation is performed every ten iterations, the final test accuracy is based on the model checkpoint that has the best validation accuracy. When there is only one training sample available, each sample is augmented 5 times using the illumination model stated in equation~\ref{eq:imageSpace}. For all the experiments we use an Adam optimizer~\cite{kingma2014adam} with a learning rate of 0.001.

\subsection{Results}
\label{sec:result}

\begin{figure}
    \centering
    \includegraphics[width=0.49\textwidth]{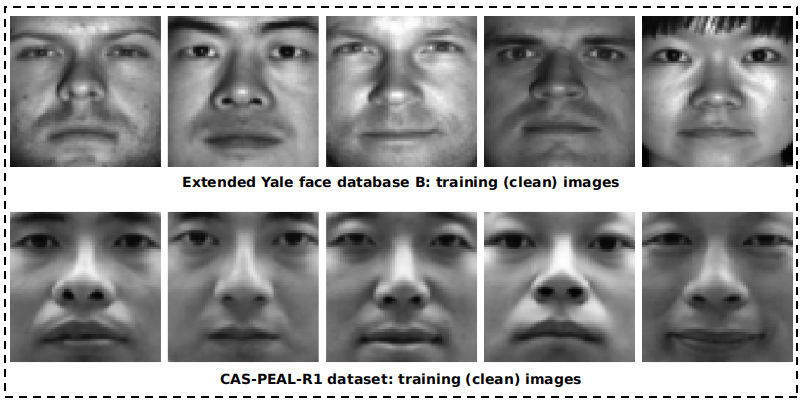}
    \caption{Clean image examples from multiple subjects: (1) Extended Yale face database B (upper row); (2) CAS-PEAL-R1 dataset (bottom row).}
    \label{fig:yale-cas-training}
\end{figure}

\begin{figure}
    \centering
    \includegraphics[width=0.49\textwidth]{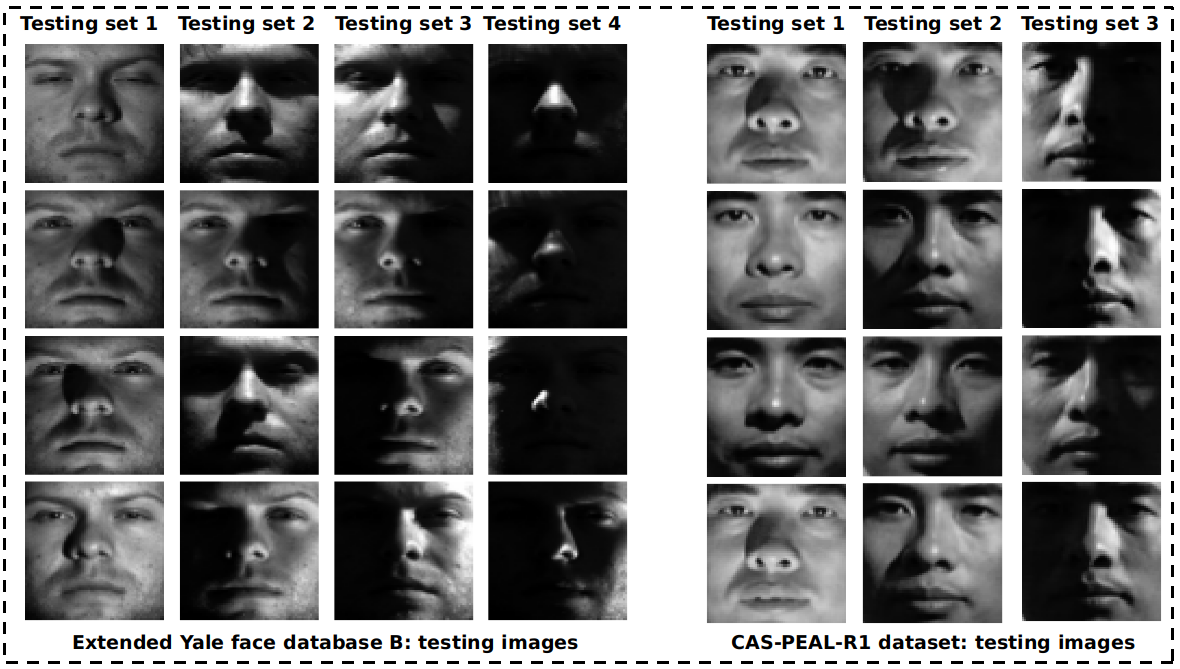}
    \caption{Image examples with changing illumination effects from one subject: (1) Extended Yale face database B (left  panel); (2) CAS-PEAL-R1 dataset (right panel).}
    \label{fig:yale-cas-testing}
\end{figure}

\begin{table*}
\centering
\caption{}
\label{tab:yaleFace}
\resizebox{0.9\textwidth}{!}{%
\begin{tabular}{|c|cccccccc|}
\hline
\multirow{3}{*}{} &
  \multicolumn{8}{c|}{\begin{tabular}[c]{@{}c@{}}Extended Yale face database\\ (Number of classes = 38)\end{tabular}} \\ \cline{2-9} 
 &
  \multicolumn{4}{c|}{\textit{Test 1}} &
  \multicolumn{4}{c|}{\textit{Test 2}} \\ \cline{2-9} 
 &
  \multicolumn{1}{c|}{\textit{Test subset 1}} &
  \multicolumn{1}{c|}{\textit{Test subset 2}} &
  \multicolumn{1}{c|}{\textit{Test subset 3}} &
  \multicolumn{1}{c|}{\textit{Test subset 4}} &
  \multicolumn{1}{c|}{\textit{Test subset 1}} &
  \multicolumn{1}{c|}{\textit{Test subset 2}} &
  \multicolumn{1}{c|}{\textit{Test subset 3}} &
  \textit{Test subset 4} \\ \hline
WebberFace + SVM &
  \multicolumn{1}{c|}{\textbf{100}$\%$} &
  \multicolumn{1}{c|}{95.1$\%$} &
  \multicolumn{1}{c|}{94.2$\%$} &
  \multicolumn{1}{c|}{90.6$\%$} &
  \multicolumn{1}{c|}{92.1$\%$} &
  \multicolumn{1}{c|}{78.8$\%$} &
  \multicolumn{1}{c|}{70.1$\%$} &
  74.4$\%$ \\ \hline
MSLDE + SVM &
  \multicolumn{1}{c|}{\textbf{100}$\%$} &
  \multicolumn{1}{c|}{93.3$\%$} &
  \multicolumn{1}{c|}{82.0$\%$} &
  \multicolumn{1}{c|}{79.1$\%$} &
  \multicolumn{1}{c|}{87.6$\%$} &
  \multicolumn{1}{c|}{64.9$\%$} &
  \multicolumn{1}{c|}{60.8$\%$} &
  64.0$\%$ \\ \hline
GradientFace + SVM &
  \multicolumn{1}{c|}{\textbf{100}$\%$} &
  \multicolumn{1}{c|}{91.5$\%$} &
  \multicolumn{1}{c|}{88.4$\%$} &
  \multicolumn{1}{c|}{85.1$\%$} &
  \multicolumn{1}{c|}{80.5$\%$} &
  \multicolumn{1}{c|}{58.6$\%$} &
  \multicolumn{1}{c|}{65.2$\%$} &
  60.0$\%$ \\ \hline
Log + DCTface + SVM &
  \multicolumn{1}{c|}{97.8$\%$} &
  \multicolumn{1}{c|}{80.5$\%$} &
  \multicolumn{1}{c|}{70.7$\%$} &
  \multicolumn{1}{c|}{43.6$\%$} &
  \multicolumn{1}{c|}{72.8$\%$} &
  \multicolumn{1}{c|}{41.3$\%$} &
  \multicolumn{1}{c|}{39.8$\%$} &
  24.6$\%$ \\ \hline
HoG + SVM &
  \multicolumn{1}{c|}{\textbf{100}$\%$} &
  \multicolumn{1}{c|}{82.4$\%$} &
  \multicolumn{1}{c|}{49.1$\%$} &
  \multicolumn{1}{c|}{65.9$\%$} &
  \multicolumn{1}{c|}{68.6$\%$} &
  \multicolumn{1}{c|}{47.2$\%$} &
  \multicolumn{1}{c|}{30.7$\%$} &
  42.0$\%$ \\ \hline
Log + HoG + SVM &
  \multicolumn{1}{c|}{\textbf{100}$\%$} &
  \multicolumn{1}{c|}{91.1$\%$} &
  \multicolumn{1}{c|}{70.5$\%$} &
  \multicolumn{1}{c|}{80.6$\%$} &
  \multicolumn{1}{c|}{73.1$\%$} &
  \multicolumn{1}{c|}{51.9$\%$} &
  \multicolumn{1}{c|}{44.2$\%$} &
  51.0$\%$ \\ \hline
VGGface &
  \multicolumn{1}{c|}{67.8$\%$} &
  \multicolumn{1}{c|}{14.6$\%$} &
  \multicolumn{1}{c|}{3.7$\%$} &
  \multicolumn{1}{c|}{2.4$\%$} &
  \multicolumn{1}{c|}{15.0$\%$} &
  \multicolumn{1}{c|}{5.7$\%$} &
  \multicolumn{1}{c|}{2.4$\%$} &
  2.6$\%$ \\ \hline
VGG face with data aug. &
  \multicolumn{1}{c|}{98.6$\%$} &
  \multicolumn{1}{c|}{77.1$\%$} &
  \multicolumn{1}{c|}{55.5$\%$} &
  \multicolumn{1}{c|}{25.3$\%$} &
  \multicolumn{1}{c|}{66.0$\%$} &
  \multicolumn{1}{c|}{33.5$\%$} &
  \multicolumn{1}{c|}{10.6$\%$} &
  7.7$\%$ \\ \hline
ResNet-50 &
  \multicolumn{1}{c|}{96.5$\%$} &
  \multicolumn{1}{c|}{50.3$\%$} &
  \multicolumn{1}{c|}{9.5$\%$} &
  \multicolumn{1}{c|}{4.0$\%$} &
  \multicolumn{1}{c|}{63.9$\%$} &
  \multicolumn{1}{c|}{14.9$\%$} &
  \multicolumn{1}{c|}{3.5$\%$} &
  2.2$\%$ \\ \hline
ResNet-50 with data aug. &
  \multicolumn{1}{c|}{97.8$\%$} &
  \multicolumn{1}{c|}{73.3$\%$} &
  \multicolumn{1}{c|}{42.0$\%$} &
  \multicolumn{1}{c|}{26.9$\%$} &
  \multicolumn{1}{c|}{91.0$\%$} &
  \multicolumn{1}{c|}{49.3$\%$} &
  \multicolumn{1}{c|}{23.8$\%$} &
  14.4$\%$ \\ \hline
DenseNet-121 &
  \multicolumn{1}{c|}{85.5$\%$} &
  \multicolumn{1}{c|}{36.4$\%$} &
  \multicolumn{1}{c|}{5.9$\%$} &
  \multicolumn{1}{c|}{3.8$\%$} &
  \multicolumn{1}{c|}{57.1$\%$} &
  \multicolumn{1}{c|}{15.5$\%$} &
  \multicolumn{1}{c|}{4.6$\%$} &
  2.0$\%$ \\ \hline
DenseNet-121 with data aug. &
  \multicolumn{1}{c|}{98.1$\%$} &
  \multicolumn{1}{c|}{58.7$\%$} &
  \multicolumn{1}{c|}{23.4$\%$} &
  \multicolumn{1}{c|}{14.8$\%$} &
  \multicolumn{1}{c|}{78.4$\%$} &
  \multicolumn{1}{c|}{24.8$\%$} &
  \multicolumn{1}{c|}{9.2$\%$} &
  2.0$\%$ \\ \hline

\begin{tabular}[c]{@{}c@{}}$\nOurMethod$\\\end{tabular} &
  \multicolumn{1}{c|}{\textbf{100}$\%$} &
  \multicolumn{1}{c|}{\textbf{98.8}$\%$} &
  \multicolumn{1}{c|}{\textbf{96.2}$\%$} &
  \multicolumn{1}{c|}{\textbf{94.4}$\%$} &
  \multicolumn{1}{c|}{\textbf{98.4}$\%$} &
  \multicolumn{1}{c|}{\textbf{95.5}$\%$} &
  \multicolumn{1}{c|}{\textbf{92.4}$\%$} &
  \textbf{91.8}$\%$ \\ \hline

\end{tabular}%
}
\label{tab:yale-results}
\end{table*}

\subsubsection{Evaluation results on extended Yale face database B} 
\label{exp.face} 

The Extended Yale Face Database B contains 38 subjects, each of which has 64 images under different illumination conditions\footnote{Please note that several subjects do not have 64 images due to corrupted images during the acquisition phase as indicated in~\cite{georghiades2001few}.}. More specifically, during the data collection stage, face images with various degrees of illumination effects are acquired by changing angles between light source direction and the camera axis (e.g., in terms of azimuth and elevation angles) as shown in Table~\ref{tab:yale-dataset-split}. Based on the degree of lighting conditions, for each subject, we split the dataset into five subsets as commonly done~\cite{zhang2009face,wang2011illumination,lai2015multiscale}: a \textit{training subset} that has 11 images, a \textit{test subset 1} that has 10 images, a \textit{test subset 2} that has 18 images, a \textit{test subset 3} that has 12 images, and a \textit{test subset 4} that has 13 images. Each image has a distinct lighting condition. The top row of Fig.~\ref{fig:yale-cas-training} shows images in the \textit{training subset} for multiple subjects, while the left panel of Fig.~\ref{fig:yale-cas-testing} exemplifies several testing samples from \textit{test subset 1} to \textit{test subset 4}. One can observe that the \textit{training set} contains images with few illumination effects, while testing sets include images with many types of lighting conditions ranging from mild to severe. The most extreme illumination condition exists in \textit{testing set 4} where subjects are hardly visible in dark environments. Table~\ref{tab:yale-dataset-split} summarizes the detailed information regarding each train and test subset.

\begin{table}
\label{tab:yaleSetup}
\centering
\caption{}
\resizebox{0.48\textwidth}{!}{%
\begin{tabular}{|c|c|c|}
\hline
              & Number of Images & Azi. \& Ele. angles                                              \\ \hline
\textit{Training set}  & 11               & -10 $\le$ \text{Azi.} $\le$ 10 \text{~and~} -20$\le$ \text{Ele.} $\le$ 20 \\ \hline
\textit{Test subset 1} & 10               & -25 $\le$ \text{Azi.} $<$-10 \text{~or~} 10$<$\text{Azi.} $\le$ 25       \\ \hline
\textit{Test subset 2} &
  18 &
  \begin{tabular}[c]{@{}c@{}}-60 $\le$ \text{Azi.} $<$-25 \text{~or~} 25$<$ \text{Azi.} $\le$ 60;\\ \text{Azi.}=0 \text{~and~} \text{Ele.} = 35 \text{~or~}= 45\end{tabular} \\ \hline
\textit{Test subset 3} & 12               & -95$\le$ \text{Azi.} $<$ -60 \text{~or~} 60 $<$ \text{Azi.} $\le$ 95       \\ \hline
\textit{Test subset 4} &
  13 &
  \begin{tabular}[c]{@{}c@{}}\text{Azi.} $<$ -95 \text{~or~} \text{Azi.}  $>$ 95;\\ \text{Azi.}=0 \text{~and~} \text{Ele.}=90\end{tabular} \\ \hline
\end{tabular}}
\label{tab:yale-dataset-split}
\end{table}

Two experiments, denoted as \textit{Test 1} and \textit{Test 2}, were conducted to evaluate the proposed approach using different training strategies for this dataset. \textit{Test 1} trains the proposed model by utilizing all images in the \textit{training set}, and performs evaluations on \textit{Test subsets 1, 2, 3, \& 4}. respectively. We further remark that in \textit{Test 1}, since there are multiple training images available, we set aside a validation set to choose the number of components in the orthogonalization step (please see Section~\ref{sec-ns}) using PCA. Specifically, We use a 66\% and 33\% of original training data for training and validation. We use the validation set to define the smallest number of components that maximize classification accuracy. \textit{Test 2} trains the proposed model using one single sample, which is randomly selected from the \textit{training set}, then it follows the same testing procedure as discussed above in \textit{Test 1}. Table~\ref{tab:yale-results} presents experimental evaluation results of \textit{Test 1} and \textit{Test 2}. Clearly, the proposed method achieves the top and most robust performance across different test subsets in both of \textit{Test 1} and \textit{Test 2}. Specifically, in \textit{Test subset 1\&2} of \textit{Test 1}, most of the approaches achieve satisfactory performance, meaning that they are able to address mild illumination effects for face recognition tasks. With increasing levels of illumination conditions in \textit{Test subset 2,3,\&4}, such as when uneven and large areas of shadow are present, other methods experienced significant decreases in performance. A similar trend can be observed in \textit{Test 2} as well. Of the comparison methods, the WebberFace, GradientFace, and MSLDE approaches consistently achieve better performance than their deep learning counterparts, indicating that they are able to eliminate some illumination effects after the "normalization" stage, especially in the test subsets (e.g., \textit{Test subset 1\&2}) with low-level of illumination variations. In addition, incorporating the log transform into HoG + SVM pipeline also improves performance.

With regards to the deep learning-based methods, some achieve excellent performance in \textit{Test subset 1\&2} of \textit{Test 1} and \textit{Test 2}, where the minimum illumination effects exist. Nevertheless, for \textit{Test subset 3\&4}, their classification accuracy decrease significantly. This may be due to the fact that only a limited number of training samples are available. As commonly done in the machine learning literature, we employed a data augmentation strategy to address the shortage of training data. Specifically, we adopt the model stated in equation~\eqref{eq:imageSpace} to randomly augment training samples with different levels of illumination (e.g., to simulate various contrast and brightness changes)\footnote{Please note in here equation~\eqref{eq:imageSpace} is applied to the entire image domain rather than a single patch.}. Each parameter in equation~\eqref{eq:imageSpace} is randomly configured as $\alpha \in [0.1, 3]$, $\beta \in [1, 30]$, $\bb \in [0.1, 3]$). We observe that for deep learning approaches illumination data augmentation strategy improves classification performance to a degree in cases where illumination changes are small (e.g., in \textit{Test subset 1\&2}). However, when illumination effects become severe, e.g., in \textit{Test subset 3\&4}, data augmentation become less effective. Though using data augmentation is a valid option to increase performance, it is still extremely difficulty to prescribe how to augment the data, in addition to other issues such as computational complexity and out of distribution performance issues ~\cite{Shifat2022invariance}. Finally, we note that the performance of deep learning-based approaches can vary significantly due to the network architecture.\\

\begin{table}
\centering
\caption{}
\label{tab:CAS-PEAL}
\resizebox{0.48\textwidth}{!}{%
\begin{tabular}{|c|ccc|}
\hline
\multirow{3}{*}{} &
  \multicolumn{3}{c|}{\begin{tabular}[c]{@{}c@{}}CAS-PEAL face dataset\\ (Number of classes = 233)\end{tabular}} \\ \cline{2-4} 
 &
  \multicolumn{1}{c|}{\textit{Test subset 1}} &
  \multicolumn{1}{c|}{\textit{Test subset 2}} &
  \textit{Test subset 3} \\ \hline
WebberFace + SVM &
  \multicolumn{1}{c|}{16.2$\%$} &
  \multicolumn{1}{c|}{19.5$\%$} &
  27.5$\%$ \\ \hline
MSLDE + SVM &
  \multicolumn{1}{c|}{18.0$\%$} &
  \multicolumn{1}{c|}{17.9$\%$} &
  40.0$\%$ \\ \hline
GradientFace + SVM &
  \multicolumn{1}{c|}{21.1$\%$} &
  \multicolumn{1}{c|}{20.0$\%$} &
  38.4$\%$ \\ \hline
Log + DCTface + SVM &
  \multicolumn{1}{c|}{11.4$\%$} &
  \multicolumn{1}{c|}{8.5$\%$} &
  4.0$\%$ \\ \hline
HoG + SVM &
  \multicolumn{1}{c|}{19.3$\%$} &
  \multicolumn{1}{c|}{13.2$\%$} &
  19.0$\%$ \\ \hline
Log + HoG + SVM &
  \multicolumn{1}{c|}{22.8$\%$} &
  \multicolumn{1}{c|}{17.2$\%$} &
  24.3$\%$ \\ \hline
VGGface &
  \multicolumn{1}{c|}{2.3$\%$} &
  \multicolumn{1}{c|}{1.7$\%$} &
  0.7$\%$ \\ \hline
VGG face with data aug. &
  \multicolumn{1}{c|}{5.5$\%$} &
  \multicolumn{1}{c|}{3.5$\%$} &
  2.6$\%$ \\ \hline
ResNet-50 &
  \multicolumn{1}{c|}{3.8$\%$} &
  \multicolumn{1}{c|}{1.8$\%$} &
  1.0$\%$ \\ \hline
ResNet-50 with data aug. &
  \multicolumn{1}{c|}{6.9$\%$} &
  \multicolumn{1}{c|}{4.8$\%$} &
  5.1$\%$ \\ \hline
DenseNet-121 &
  \multicolumn{1}{c|}{1.4$\%$} &
  \multicolumn{1}{c|}{1.1$\%$} &
  0.8$\%$ \\ \hline
DenseNet-121 with data aug. &
  \multicolumn{1}{c|}{3.3$\%$} &
  \multicolumn{1}{c|}{2.0$\%$} &
  1.0$\%$ \\ \hline

$\nOurMethod$ &
  \multicolumn{1}{c|}{\textbf{42.2}$\%$} &
  \multicolumn{1}{c|}{\textbf{50.2}$\%$} &
  \textbf{49.7}$\%$ \\ \hline
\end{tabular}%
}
\label{tab:cas-peal-results}
\end{table}

\subsubsection{Evaluation results on CAS-PEAL-R1 dataset} CAS-PEAL-R1 is a face recognition dataset that contains 233 different subjects. Within this dataset, for each subject there is only one "clean" (with standard illumination) image and at least 9 images with varying lighting effects. The azimuth angle and elevation angle between light source direction and the camera axis were selected to be $\{-90, -45, 0, 45, 90\}$ and $\{-45, 0, 45\}$, respectively. Based on the azimuth angle, we divide the face images with various levels of illumination into three different subsets: \textit{Test subset 1} where azimuth angle equals to 0 degree, \textit{Test subset 2} where azimuth angle is either -45 or 45 degree, and \textit{Test subset 3} where azimuth angle is either -90 or 90 degree. Figures~\ref{fig:yale-cas-training} and \ref{fig:yale-cas-testing} illustrate sample images from the CAS-PEAL-R1 datset. Specifically, the bottom row of Fig.~\ref{fig:yale-cas-training} shows "clean" images for multiple subjects. The right panel of Fig.~\ref{fig:yale-cas-testing} shows images with changing lighting conditions of one subject. In this experiment, the model only utilizes the single clean image of each subject for training. Evaluations are performed on three testing subsets: \textit{Test subset 1, 2, \& 3}. Table~\ref{tab:cas-peal-results} summarizes the experiment results. Overall, the proposed method outperforms other alternatives by a large margin, which is a consistent trend across the three testing subsets. With respect to the comparison methods, GradientFace and MSLDE achieve the best performance. Deep learning-based approaches are not able to perform well perhaps due to the limited number of training samples, even illumination data augmentation strategies are used. \\

\begin{figure}
    \centering
    \includegraphics[width=0.49\textwidth]{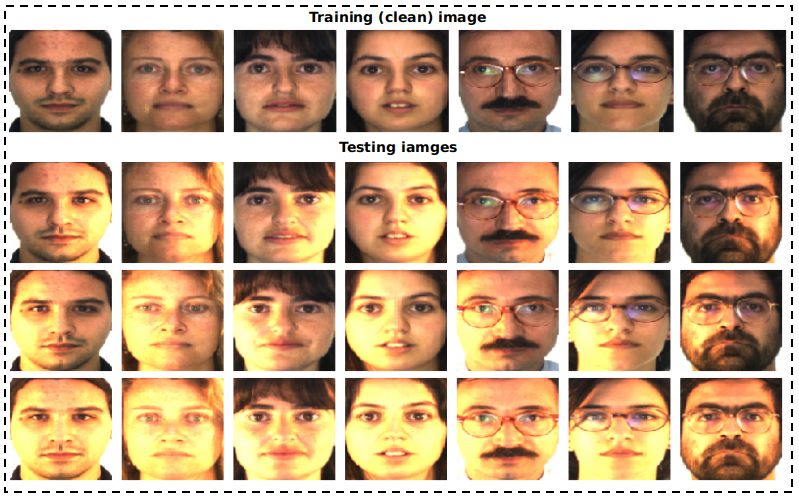}
    \caption{Image examples from AR face dataset: (1) Top row shows clean images from multiple subjects; (2) second to fourth row show images with illumination variations (e.g., right light on, left light on, and both sides light on).}
    \label{fig:arface-train-test}
\end{figure}

\subsubsection{Evaluation results on AR face dataset} Finally, we evaluate and compare the proposed method on AR face dataset which has 100 subjects. Compared with the previous two datasets, the AR face dataset is a less challenging dataset as it has mild illumination variations as shown in Fig.~\ref{fig:arface-train-test}. More specifically, each subject has two sets of images. Each set contains one "clean" image and 3 images with different levels of illumination. In total, each subject has 2 clean images and 6 images with different lighting variations (e.g., 2 images with left light on, 2 images with right light on, and 4 images with all side lights on). We combine images with left and right illumination on as the \textit{test subset 1}, while images with side lighting on are included for \textit{test subset 2}. Similarly, each model is trained using 2 clean images, and is tested on \textit{test subset 1} and \textit{test subset 2} respectively. Table~\ref{tab:arface-result} provides the classification accuracy for \textit{test subset 1 \&2}. We notice that the method Log + HoG~\cite{zhu2013logarithm} method achieves the best performance while the method proposed here obtains comparable performance. Other methods such as MSLDE and GradientFace also have high accuracies. It is worth noting that illumination data augmentation are most effective for the AR face dataset, which allows deep learning approaches to gain significant performance improvement, implying that employing data augmentation is an effective route to address mild illumination variations.

\begin{table}
\centering
\caption{}
\label{tab:Arface}
\resizebox{0.45\textwidth}{!}{%
\begin{tabular}{|c|cc|}
\hline
\multirow{3}{*}{} &
  \multicolumn{2}{c|}{\begin{tabular}[c]{@{}c@{}}ARface dataset\\ (Number of classes = 100)\end{tabular}} \\ \cline{2-3} 
 &
  \multicolumn{1}{c|}{\textit{Test subset 1}} &
  \textit{Test subset 2} \\ \hline
WebberFace + SVM &
  \multicolumn{1}{c|}{93.7$\%$} &
  87.0$\%$ \\ \hline
MSLDE + SVM &
  \multicolumn{1}{c|}{97.5$\%$} &
  94.5$\%$ \\ \hline
GradientFace + SVM &
  \multicolumn{1}{c|}{97.5$\%$} &
  93.0$\%$ \\ \hline
Log + DCTface + SVM &
  \multicolumn{1}{c|}{87.5$\%$} &
  73.0$\%$ \\ \hline
HoG + SVM &
  \multicolumn{1}{c|}{99.7$\%$} &
  83.5$\%$ \\ \hline
Log + HoG + SVM &
  \multicolumn{1}{c|}{\textbf{100.0}$\%$} &
  \textbf{95.5}$\%$ \\ \hline
VGGface &
  \multicolumn{1}{c|}{19.7$\%$} &
  11.5$\%$ \\ \hline
VGG face with data aug. &
  \multicolumn{1}{c|}{80.2$\%$} &
  58.4$\%$ \\ \hline
ResNet-50 &
  \multicolumn{1}{c|}{59.5$\%$} &
  7.5$\%$ \\ \hline
ResNet-50 with data aug. &
  \multicolumn{1}{c|}{92.7$\%$} &
  80.0$\%$ \\ \hline
DenseNet-121 &
  \multicolumn{1}{c|}{54.2$\%$} &
  9.0$\%$ \\ \hline
DenseNet-121 with data aug. &
  \multicolumn{1}{c|}{89.9$\%$} &
  82.4$\%$ \\ \hline
\begin{tabular}[c]{@{}c@{}}$\nOurMethod$\end{tabular} &
  \multicolumn{1}{c|}{99.5$\%$} &
  94.5$\%$ \\ \hline
\end{tabular}%
}
\label{tab:arface-result}
\end{table}
\section{Discussion}
\label{sec-dis}

The experimental results presented above show that the face recognition approach we proposed is more robust to variations in illumination conditions than a variety of existing methods. The method is based on splitting face images into finite support neighborhoods $\Omega_k$, and using a Discrete R-CDT representation for the gradient distribution within each neighborhood. As such, certain parametric choices (number of projections in R-CDT, neighborhood size, and neighborhood overlap) have to be made. To study the best choices for these parameters, we followed the same training and testing strategy used in \textit{Test 2} for the Yale Face dataset and focus on the \textit{Test subset 4}, which has the most challenging illumination conditions. Results are present in the Table~\ref{tab:hyperparameter} in Section.\ref{sec:tables} (appendix). When varying different patch sizes, the overlap size is set as 0. Overall, results show that utilizing smaller patch improves the performance. We postulate this may be because the illumination model described in equation \eqref{eq:imageSpace} is more accurate for small neighborhoods. Secondly, when the neighborhood size is fixed, increasing number of projections and overlap size are able to provide performance improvement to some extend. 

\subsection{Future work}
\label{sec:futurework}

The proposed approach above builds on prior work related to patch-wise analysis of gradient distributions. One of the novel contributions described in our method is the modeling of certain illumination transformations as transport operations on the resulting gradient distributions. Based on this point of view, there are multiple interesting perspectives worth further exploring as future work. Firstly, as discussed in Section~\ref{sec:problemIntro}, the log pixel transformation ~\cite{gonzalez2002digital} is a widely used technique to circumvent illumination effects as shown in prior work~\cite{chen2004illumination, zhu2013logarithm,lai2015multiscale}. Based on this we also applied the log transform as a preproccessing step for our method. Table~\ref{tab:log-performance} in Section \ref{sec:tables} (appendix) demonstrates classification results as well as the performance improvement (as highlighted using a upward arrow) for three face datasets used in the experiment. In general, employing the log transform increases classification accuracy of our method for most  tests. In addition, the largest performance increase takes place in where the dataset has large regions of illumination variations such as \textit{test subset 4} in the Yale Face dataset. We note that the application of the log transform on image space changes equation~\eqref{eq:imageSpace}. Therefore, a careful mathematical analysis of incorporating the log transform in our gradient distribution representation will be the subject of future work.

Secondly, the affine illumination generative model in equation~\eqref{eq:affineModel} includes spatial transformations such as scaling and translation of the 2D discrete gradient distribution. Other 2D discrete gradient distribution deformations such as rotation, anisotropic scaling, and shearing, are worth exploring as future work in connection with more elaborate illumination models. Specifically, Shifat-E-Rabbi $et. al.$~\cite{Shifat2022invariance} developed a mathematical formulation for rotation, anisotropic scaling, and shear deformations in the sliced-Wasserstein space. Future work will consider adopting the same idea to model rotation, anisotropic scaling, and shearing effects of 2D discrete gradient distribution in discrete R-CDT domain, if more complex illumination models require it.

We also note that when recognizing faces certain image patches contain more discriminative information than others. For example patches near the eyes, nose, and mouth, tend to contain more information humans use to identify one another. In this pilot study, when aggregating the distance $d^c_k$ from each image patch in equation~\eqref{eq.dis}, we implicitly assume that each image patch has a equal weight coefficient. However, we can rewrite the equation~\eqref{eq.dis} $\text{argmin}_c \sum_{k=1}^K d^c_k$ as $\text{argmin}_c \sum_{k=1}^K p_kd^c_k$, where $p_k$ is the weight coefficient of $k_\text{th}$ patch. This means that certain $p_k$s will have higher weight than others. Therefore, future study will investigate a learning-based approach to derive such weight coefficients $p_k$s for potential performance improvements.

Finally, in equation~\eqref{eq.subspace} we argue that taking the span of multiple transformed training examples allows the proposed model to capture more complicated lighting variations than the ones captured by the set $\mathbb{H}_0$. Table~\ref{tab:yale-results} demonstrates this by showing that an increasing number of training images, the classification accuracy improves as more data is added for each test subset. Furthermore, we conducted a similar experiment on the CAS-PEAL-R1 face dataset. Note that, originally, the CAS-PEAL-R1 face dataset only has one "clean" image for each subject. In order to expand the training set, we added images from \textit{test subset 1} and \textit{test subset 2} to the training set and evaluated the algorithm on \textit{test subset 3}. Results shows that the classification accuracy increases from 49.7\% to 70.17\% when training images include \textit{test subset 1} images, and to 82.51\% when images from both of \textit{test subset 1} and \textit{test subset 2} are included for training, respectively. These results indicate the proposed method is learning gradient distribution transformations that go beyond $\mathbb{H}_0$ (scaling and translation). As such, we believe that enhanced methods for combining data (in transform domain) and illumination models $\mathbb{H}$ could potentially yield even higher accuracy results.

\section{Conclusion}

We proposed a novel transport-based approach for illumination-invariant face recognition. We first showed that changing illumination conditions contribute to certain types of deformations of local 2D discrete gradient distributions in local image patches. Then we mathematically showed that the patch-wise image gradient distributions observed under certain illumination variations form a convex set in the Discrete R-CDT domain, and can thus be separated using a nearest subspace method. Experiment results demonstrated superior performance of the proposed method in multiple face datasets with illumination variations in challenging lighting conditions. Finally, we provided a detailed discussion regarding potential approaches for increasing the performance of the method even further. 

\section*{Acknowledgments}
This work was supported in part by NIH award GM130825.

\bibliographystyle{IEEEbib}
\bibliography{ref}

\begin{thebibliography}{10}

\bibitem{kolouri2015radon}
Soheil Kolouri, Se~Rim Park, and Gustavo~K Rohde,
\newblock ``The radon cumulative distribution transform and its application to
  image classification,''
\newblock {\em IEEE transactions on image processing}, vol. 25, no. 2, pp.
  920--934, 2015.

\bibitem{software}
Yan Zhuang, Shiying Li, Mohammad Shifat-E-Rabbi, Xuwang Yin, Abu
  Hasnat~Mohammad Rubaiyat, and Gustavo~Kunde Rohde,
\newblock ``Python implementation of discrete rcdt for illumination-invariant
  face recognition,'' https://github.com/rohdelab/drcdt\_face.

\bibitem{package}
Imaging Data~Science Laboratory,
\newblock ``Pytranskit,'' https://github.com/rohdelab/PyTransKit.

\bibitem{belhumeur1997eigenfaces}
Peter~N. Belhumeur, Jo{\~a}o~P Hespanha, and David~J. Kriegman,
\newblock ``Eigenfaces vs. fisherfaces: Recognition using class specific linear
  projection,''
\newblock {\em IEEE Transactions on pattern analysis and machine intelligence},
  vol. 19, no. 7, pp. 711--720, 1997.

\bibitem{adini1997face}
Yael Adini, Yael Moses, and Shimon Ullman,
\newblock ``Face recognition: The problem of compensating for changes in
  illumination direction,''
\newblock {\em IEEE Transactions on pattern analysis and machine intelligence},
  vol. 19, no. 7, pp. 721--732, 1997.

\bibitem{georghiades2001few}
Athinodoros~S. Georghiades, Peter~N. Belhumeur, and David~J. Kriegman,
\newblock ``From few to many: Illumination cone models for face recognition
  under variable lighting and pose,''
\newblock {\em IEEE transactions on pattern analysis and machine intelligence},
  vol. 23, no. 6, pp. 643--660, 2001.

\bibitem{lai2015multiscale}
Zhao-Rong Lai, Dao-Qing Dai, Chuan-Xian Ren, and Ke-Kun Huang,
\newblock ``Multiscale logarithm difference edgemaps for face recognition
  against varying lighting conditions,''
\newblock {\em IEEE transactions on image processing}, vol. 24, no. 6, pp.
  1735--1747, 2015.

\bibitem{chen2004illumination}
Weilong Chen, Meng~Joo Er, and Shiqian Wu,
\newblock ``Illumination compensation and normalization using logarithm and
  discrete cosine transform,''
\newblock in {\em ICARCV 2004 8th Control, Automation, Robotics and Vision
  Conference, 2004.} IEEE, 2004, vol.~1, pp. 380--385.

\bibitem{zhang2009face}
Taiping Zhang, Yuan~Yan Tang, Bin Fang, Zhaowei Shang, and Xiaoyu Liu,
\newblock ``Face recognition under varying illumination using gradientfaces,''
\newblock {\em IEEE Transactions on image processing}, vol. 18, no. 11, pp.
  2599--2606, 2009.

\bibitem{wang2011illumination}
Biao Wang, Weifeng Li, Wenming Yang, and Qingmin Liao,
\newblock ``Illumination normalization based on weber's law with application to
  face recognition,''
\newblock {\em IEEE Signal Processing Letters}, vol. 18, no. 8, pp. 462--465,
  2011.

\bibitem{qawaqneh2017deep}
Zakariya Qawaqneh, Arafat~Abu Mallouh, and Buket~D Barkana,
\newblock ``Deep convolutional neural network for age estimation based on
  vgg-face model,''
\newblock {\em arXiv preprint arXiv:1709.01664}, 2017.

\bibitem{he2016deep}
Kaiming He, Xiangyu Zhang, Shaoqing Ren, and Jian Sun,
\newblock ``Deep residual learning for image recognition,''
\newblock in {\em Proceedings of the IEEE conference on computer vision and
  pattern recognition}, 2016, pp. 770--778.

\bibitem{simonyan2014very}
Karen Simonyan and Andrew Zisserman,
\newblock ``Very deep convolutional networks for large-scale image
  recognition,''
\newblock {\em arXiv preprint arXiv:1409.1556}, 2014.

\bibitem{huang2017densely}
Gao Huang, Zhuang Liu, Laurens Van Der~Maaten, and Kilian~Q Weinberger,
\newblock ``Densely connected convolutional networks,''
\newblock in {\em Proceedings of the IEEE conference on computer vision and
  pattern recognition}, 2017, pp. 4700--4708.

\bibitem{dalal2005histograms}
Navneet Dalal and Bill Triggs,
\newblock ``Histograms of oriented gradients for human detection,''
\newblock in {\em 2005 IEEE computer society conference on computer vision and
  pattern recognition (CVPR'05)}. IEEE, 2005, vol.~1, pp. 886--893.

\bibitem{zhu2013logarithm}
Jun-Yong Zhu, Wei-Shi Zheng, and Jian-Huang Lai,
\newblock ``Logarithm gradient histogram: A general illumination invariant
  descriptor for face recognition,''
\newblock in {\em 2013 10th IEEE International Conference and Workshops on
  Automatic Face and Gesture Recognition (FG)}. IEEE, 2013, pp. 1--8.

\bibitem{basri2003lambertian}
Ronen Basri and David~W Jacobs,
\newblock ``Lambertian reflectance and linear subspaces,''
\newblock {\em IEEE transactions on pattern analysis and machine intelligence},
  vol. 25, no. 2, pp. 218--233, 2003.

\bibitem{ho2005effect}
Jeffrey Ho and David Kriegman,
\newblock ``On the effect of illumination and face recognition,''
\newblock {\em Face Processing: Advanced Modeling and Methods}, 2005.

\bibitem{wright2008robust}
John Wright, Allen~Y Yang, Arvind Ganesh, S~Shankar Sastry, and Yi~Ma,
\newblock ``Robust face recognition via sparse representation,''
\newblock {\em IEEE transactions on pattern analysis and machine intelligence},
  vol. 31, no. 2, pp. 210--227, 2008.

\bibitem{parkhi2015deep}
Omkar~M. Parkhi, Andrea Vedaldi, and Andrew Zisserman,
\newblock ``Deep face recognition,''
\newblock in {\em Proceedings of the British Machine Vision Conference (BMVC)}.
  September 2015, pp. 41.1--41.12, BMVA Press.

\bibitem{shorten2019survey}
Connor Shorten and Taghi~M Khoshgoftaar,
\newblock ``A survey on image data augmentation for deep learning,''
\newblock {\em Journal of Big Data}, vol. 6, no. 1, pp. 1--48, 2019.

\bibitem{lowe1999object}
David~G Lowe,
\newblock ``Object recognition from local scale-invariant features,''
\newblock in {\em Proceedings of the seventh IEEE international conference on
  computer vision}. Ieee, 1999, vol.~2, pp. 1150--1157.

\bibitem{felzenszwalb2009object}
Pedro~F Felzenszwalb, Ross~B Girshick, David McAllester, and Deva Ramanan,
\newblock ``Object detection with discriminatively trained part-based models,''
\newblock {\em IEEE transactions on pattern analysis and machine intelligence},
  vol. 32, no. 9, pp. 1627--1645, 2009.

\bibitem{ahonen2006face}
Timo Ahonen, Abdenour Hadid, and Matti Pietikainen,
\newblock ``Face description with local binary patterns: Application to face
  recognition,''
\newblock {\em IEEE transactions on pattern analysis and machine intelligence},
  vol. 28, no. 12, pp. 2037--2041, 2006.

\bibitem{chen2009wld}
Jie Chen, Shiguang Shan, Chu He, Guoying Zhao, Matti Pietik{\"a}inen, Xilin
  Chen, and Wen Gao,
\newblock ``Wld: A robust local image descriptor,''
\newblock {\em IEEE transactions on pattern analysis and machine intelligence},
  vol. 32, no. 9, pp. 1705--1720, 2009.

\bibitem{sariyanidi2014automatic}
Evangelos Sariyanidi, Hatice Gunes, and Andrea Cavallaro,
\newblock ``Automatic analysis of facial affect: A survey of registration,
  representation, and recognition,''
\newblock {\em IEEE transactions on pattern analysis and machine intelligence},
  vol. 37, no. 6, pp. 1113--1133, 2014.

\bibitem{shifat2020radon}
Mohammad Shifat-E-Rabbi, Xuwang Yin, Abu Hasnat~Mohammad Rubaiyat, Shiying Li,
  Soheil Kolouri, Akram Aldroubi, Jonathan~M Nichols, Gustavo~K Rohde, et~al.,
\newblock ``Radon cumulative distribution transform subspace modeling for image
  classification,''
\newblock {\em J Math Imaging Vis}, vol. 63, pp. 1185--1203, 2021.

\bibitem{gonzalez2002digital}
Rafael~C Gonzalez, Richard~E Woods, et~al.,
\newblock ``Digital image processing,'' 2002.

\bibitem{epstein19955}
Russell Epstein, Peter Hallinan, and Alan Yuille,
\newblock ``5$\pm$2 eigenimages suffice: An empirical investigation of
  low-dimensional lighting models,''
\newblock in {\em IEEE Workshop on Physics-Based Vision}, 1995, pp. 108--116.

\bibitem{ramamoorthi2002analytic}
Ravi Ramamoorthi,
\newblock ``Analytic pca construction for theoretical analysis of lighting
  variability in images of a lambertian object,''
\newblock {\em IEEE transactions on pattern analysis and machine intelligence},
  vol. 24, no. 10, pp. 1322--1333, 2002.

\bibitem{azulay2018deep}
Aharon Azulay and Yair Weiss,
\newblock ``Why do deep convolutional networks generalize so poorly to small
  image transformations?,''
\newblock {\em arXiv preprint arXiv:1805.12177}, 2018.

\bibitem{park2018cumulative}
Se~Rim Park, Soheil Kolouri, Shinjini Kundu, and Gustavo~K Rohde,
\newblock ``The cumulative distribution transform and linear pattern
  classification,''
\newblock {\em Applied and computational harmonic analysis}, vol. 45, no. 3,
  pp. 616--641, 2018.

\bibitem{kolouri2017optimal}
Soheil Kolouri, Se~Rim Park, Matthew Thorpe, Dejan Slepcev, and Gustavo~K
  Rohde,
\newblock ``Optimal mass transport: Signal processing and machine-learning
  applications,''
\newblock {\em IEEE signal processing magazine}, vol. 34, no. 4, pp. 43--59,
  2017.

\bibitem{rabin2011wasserstein}
Julien Rabin, Gabriel Peyr{\'e}, Julie Delon, and Marc Bernot,
\newblock ``Wasserstein barycenter and its application to texture mixing,''
\newblock in {\em International Conference on Scale Space and Variational
  Methods in Computer Vision}. Springer, 2011, pp. 435--446.

\bibitem{santambrogio2015optimal}
Filippo Santambrogio,
\newblock ``Optimal transport for applied mathematicians,''
\newblock {\em Birk{\"a}user, NY}, vol. 55, no. 58-63, pp. 94, 2015.

\bibitem{Shifat2022invariance}
Mohammad Shifat-E-Rabbi, Yan Zhuang, Shiying Li, Abu Hasnat~Mohammad Rubaiyat,
  Xuwang Yin, Gustavo~K Rohde, et~al.,
\newblock ``Invariance encoding in sliced-wasserstein space for image
  classification with limited training data,''
\newblock {\em arXiv preprint arXiv:2201.02980}, 2022.

\bibitem{bishop2006pattern}
Christopher~M Bishop and Nasser~M Nasrabadi,
\newblock {\em Pattern recognition and machine learning}, vol.~4,
\newblock Springer, 2006.

\bibitem{martinez1998ar}
Aleix Martinez and Robert Benavente,
\newblock ``The ar face database: Cvc technical report, 24,''
\newblock Tech. {R}ep., The Ohio State University, 1998.

\bibitem{gao2007cas}
Wen Gao, Bo~Cao, Shiguang Shan, Xilin Chen, Delong Zhou, Xiaohua Zhang, and
  Debin Zhao,
\newblock ``The cas-peal large-scale chinese face database and baseline
  evaluations,''
\newblock {\em IEEE Transactions on Systems, Man, and Cybernetics-Part A:
  Systems and Humans}, vol. 38, no. 1, pp. 149--161, 2007.

\bibitem{kingma2014adam}
Diederik~P Kingma and Jimmy Ba,
\newblock ``Adam: A method for stochastic optimization,''
\newblock {\em arXiv preprint arXiv:1412.6980}, 2014.

\end{thebibliography}

\clearpage

\section{Appendix}
\subsection{Simulated illumination effects}
\label{sec:simulatedAffine}
\begin{figure}[hbt]
\centering
  \includegraphics[width=0.48\textwidth]{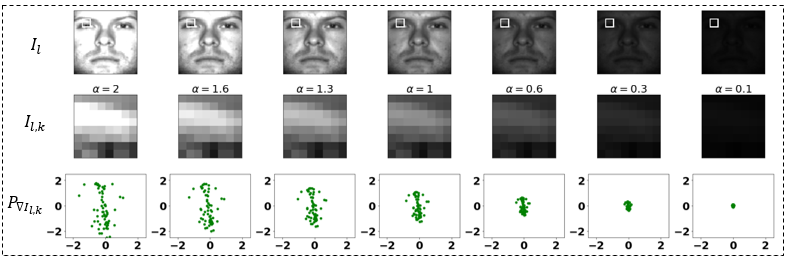}
\caption{Top and middle panels show simulated illumination conditions in the image space using equation \eqref{eq:imageSpace} where $\beta=0$ and $\bb=\textbf{0}$. The bottom panel demonstrates the corresponding scaling effects caused by different $\alpha$ values in 2D discrete distribution space.}
\label{fig:scaleGradients}
\end{figure}

\begin{figure}[hbt]
\centering
  \includegraphics[width=0.48\textwidth]{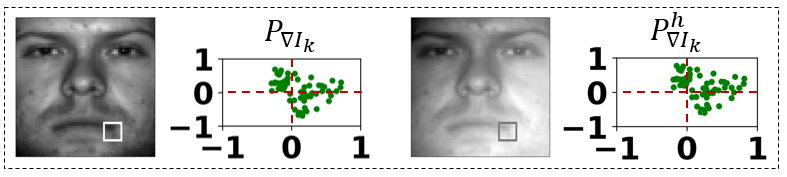}
\caption{Simulated illumination conditions in the image space using equation \eqref{eq:imageSpace} where $\alpha = 1$, $\beta = 0$, and $\bb$ is a constant vector. Adding a linear gradient is able to simulate lighting coming from the side, therefore, resulting in a translation effect in 2D discrete distribution space.}
\label{fig:linearGradients}
\end{figure}

Section \ref{sec:problemIntro} illustrates effects of varying illumination conditions on local 2D gradient distributions. To be precise, varying parameters $\alpha$ and $\bb$ of the illumination model, defined in equation \eqref{eq:h_model}, can lead to changes of illumination intensity and linear illumination gradient, as discussed in equation \eqref{eq:imageSpace}. Here Fig.\ref{fig:scaleGradients} and Fig.\ref{fig:linearGradients} provide visualizations of such variations. Note that equation~\eqref{eq:imageSpace} is applied to the entire image domain for the illustration purposes. Fig.\ref{fig:scaleGradients} shows that changing $\alpha$ not only contributes to contrast change (top two rows) in image space but also causes scaling effects for the corresponding local 2D discrete distributions (bottom row). Likewise in Fig.\ref{fig:linearGradients}, the original face image is shown on the left, we add the linear illumination gradient, specified by a constant vector $\bb$, to simulate the lighting coming from the side, as shown one in right part of Fig.\ref{fig:linearGradients}. Correspondingly, the local 2D discrete distribution of the simulated image experiences a translation effect. In addition, in equation~\eqref{eq:imageSpace}, $\beta$ is responsible for controlling image brightness. However, it is eliminated automatically in derivative operation when calculating image gradients. Please note that there are several existing works that aimed to address the contrast variations ($\alpha$ in equation~\eqref{eq:imageSpace}) by normalization. For instance, HoG features perform block-wise $L_2$-norm normalization within a local region~\cite{dalal2005histograms}. Furthermore, applying the log transform is also able to normalize the contrast, by expanding the value of dark pixels~\cite{gonzalez2002digital}. Though these approaches improve performance to certain degree but are ineffective for challenging illumination conditions as demonstrated in Section~\ref{sec:result}.

\subsection{Dirac measure}\label{append:dirac}
A Dirac measure $\delta_{\bz}$ on $\R^n$ is a measure with mass concentrated at $\bz\in\R^n$ ($n\geq1$) such that for any (measurable) set $A\subseteq \R^n$
\begin{equation}
	\delta_{\bz}(A)= \begin{cases}
		&1, \quad \bz\in A\\
		&0, \quad \bz\notin A.
	\end{cases}
\end{equation}
In other words, a set has measure $1$  if it contains the point $z$ and measure zero otherwise. In particular, $\delta_{\bz} (\{\bz\})$ =1 and $\delta_{\bz} (\{\bz^{\prime}\})$ =0 for any $\bz^{\prime}\neq \bz$.

\subsection{\transname~for 1D Discrete Distribution}
\label{sec:1d-cdt}

\subsubsection{Connection to the Wasserstein distance} 
Given two 1D discrete distributions $P_{Z^{(1)}}$ and $P_{Z^{(2)}}$ in $\mP(\R)$, 
\begin{equation}
	W_2(P_{Z^{(1)}}, P_{Z^{(2)}}) = \sqrt{\frac{1}{N}||\F(P_{Z^{(1)}})-\F(P_{Z^{(1)}})||^2},
\end{equation}
where $||\cdot||$ denotes the Euclidean distance on $\R^N$.
{Indeed, for  two discrete measures  $P_{Z^{(1)}}= \frac{1}{N}\sum\limits_{i=1}^N \delta_{z_i^{(1)}}$ and $P_{Z^{(2)}} =\frac{1}{N} \sum\limits_{i=1}^N \delta_{z_i^{(2)}}$,  the 2-Wasserstein distance between the two measures is the same as the Euclidean distance between the mass location vectors sorted in an increasing order,  i.e.,  $\wass{P_{Z^{(1)}}}{P_{Z^{(2)}}} =\sqrt{\frac{1}{N}\sum_{i=1}^N(\widetilde z_i^{(1)}-\widetilde z_i^{(2)})^2}$, where $\widetilde z_1^1\leq \cdots \leq \widetilde z_N^1$ and  $\widetilde z_1^2\leq \cdots \leq \widetilde z_N^2$ are sorted versions of $z_1^{(1)},...,z_N^{(1)}$ and $z_1^{(2)},...,z_N^{(2)}$ respectively. This can be seen as a special case of Proposition 2.17 in \cite{santambrogio2015optimal}}.

\subsubsection{Composition property}
\label{1d-cdt-composition-property}

Let $T:\R\rightarrow\R$ be a strictly increasing function and $P_Z\in \mP_N(\R) $, then 
\begin{equation}
    \F\big(P_Z^T\big) = T\circ \F\big(P_Z\big),
\label{eq.1dcdtProb}
\end{equation}
where the composition is operated entrywise and $ P_Z^T =\frac{1}{N}\sum_{i=1}^{N}\delta_{T(z_i))}$ is the push-forward  \footnote{In general the push-forward  measure $T_{\sharp}\mu$ of a measure $\mu$ under $T:X\rightarrow Y$ is defined by the property that $T_{\sharp}\mu(U)= \mu(T^{-1}(U))$ for any measurable set $U\subseteq Y$. In particular, for any one-dimensional discrete distribution $P_Z$,  given any Lebesgue measurable function $T:\R \rightarrow \R$, it can be shown that $T_{\sharp}(P_Z) = \frac{1}{N}\sum_{i=1}^N\delta_{T(z_i)}$.  } distribution of $P_Z$ by $T$. The property implies that for any strictly increasing $T$, applying the transform $\F$ to the push-forward distribution $P_Z^T$ from $P_Z$ via $T$ equates to composing T with the transform $\F(P_Z)$.

\noindent Proof: Let $P_Z= \frac{1}{N}\sum_{i=1}^{N}\delta_{z_i}$ with $  \F(P_Z)=[
                \widetilde{z}_1,  \widetilde{z}_2,  \cdots  \widetilde{z}_N]^T$ where $\widetilde z_1\leq \widetilde z_2\leq \cdots \widetilde z_N$ is reordered from $[z_1,...,z_N]$. Observing that $P_Z^T = \frac{1}{N}\sum_{i=1}^N\delta_{T(z_i)}$ and by the definition of \transname, $\F(P_Z^T)$ is the vector of which the entries are $T(z_i)$'s in an increasing order, i.e, 
                \begin{equation}
                \F(P_Z^T)=[
               \widetilde{ T(z_1)},  \widetilde{ T(z_2)},  \cdots \widetilde{ T(z_N)}]^T
       \end{equation}
            where  $\widetilde{ T(z_1)}\leq \widetilde {T(z_2)}\leq \cdots \leq \widetilde {T(z_N)}$. On the other hand, since $T$ is strictly increasing, we have that $ T(\widetilde z_1)\leq T(\widetilde z_2)\leq \cdots \leq T(\widetilde z_N)$, which is also an reordering of $T(z_1),...,T(z_N)$. Hence we have that 
            \begin{equation}
                \F(P_Z^T)=\begin{bmatrix}
                T(\widetilde{z}_1) \\ T(\widetilde{z}_2) \\ \vdots \\ T(\widetilde{z}_N)
            \end{bmatrix} = T\circ \F(P_Z).
            \end{equation}
            

We note the following interesting cases where $T$ is a translation or a scaling diffeomorphism.

\subsubsection{Translation property}
Let $T:\R\rightarrow\R$ be the translation function  where $T(x)=x+a$ for some $a\in \R$. By the composition property, we have that \begin{equation}
   \F(P_Z^T)= \F(P_Z)+a, 
\end{equation}

where the addition on the RHS is operated entry-wise.

\subsubsection{Scaling property}

Let $T:\R\rightarrow\R$ be a scaling where $T(x)= cx$ for some $c>0$. By the composition property, we have that
\begin{equation}
     \F(P_Z^T)= c\F(P_Z).
\end{equation}

This composition property can addresses several deformations, specified by $T$, for one-dimensional discrete distribution, thus rendering classes convex and simplifying the classification task.

\subsubsection{Convexity property}

\label{1dcdtconvexity}
Using the composition property, we can derive the following convexity property of the transform $\F$. Let $\mG_1\subseteq \mH_1$ where $\mH_1 = \{T:\R\rightarrow \R : T ~\text{is a strictly increasing diffeomorphism}\}$. Then given a one dimensional discrete distribution $P_Z$ for some $Z =\{ z_i\}_{i=1}^N$, the set of transforms $\F \big(\mP_Z^{\mG_1}\big):=\{\F(P_Z^T)\mid T\in \mG_1\}$ is convex if $\mG_1$ is convex. \\

\noindent Proof: Let $T_1, T_2 \in\mG$ and $\lambda \in [0,1]$. Then by the definition of the {\transname} and using the composition property of \transname, we have that
\begin{align}
  &\lambda \F \Big(P_Z^{T_1}\Big) +(1-\lambda)  \F\Big(P_Z^{T_2}\Big) \\=&  \lambda \Big(T_1\circ \F\big(P_Z\big)\Big)+(1-\lambda)  \Big(T_2\circ  \F\big(P_Z\big)\Big)
  \\ =&  \big(\lambda T_1 +(1-\lambda)T_2\big) \circ\F\big({P_Z}\big) 
\\=&  \F\Big(P^{\lambda T_1 +(1-\lambda)T_2}_Z\Big)\in \F \big(\mP_Z^{\mG_1}\big):\label{eq:inclusion}
\end{align}
where the inclusion in \eqref{eq:inclusion} is due to the fact that  $\mG_1$ is convex and in particular, $\lambda T_1 +(1-\lambda)T_2)\in \mG_1$. Hence $\F \big(\mP_Z^{\mG_1}\big):$ is convex.

\subsection{\hogname~for 2D discrete distribution}
\subsubsection{Convexity property}
\label{sec:2DConvexityProof}
The set $\F^{*}\big(\nSetHNot\big)$ is convex.

\noindent Proof: 
Let $P_{\grad I_k^c}^{h_1},P_{\grad I_k^c}^{h_2} \in \nSetHNot$ where $h_1(\bz) = a_1\bz+\bb_1, h_2(\bz) = a_2\bz+\bb_2\in \mathbb H_0$. By definition, for any $\theta\in [0,\pi)$
\begin{align*}
	\F^{*}\big(P_{\grad I_k^c}^{h_1}\big)(\theta) &= a_1 \F^{*}\big(P(\grad I_k^c)\big)+\bb_1\cdot \wtheta\\
	\F^{*}\big(P_{\grad I_k^c}^{h_2}\big)(\theta) &= a_2 \F^{*}\big(P(\grad I_k^c)\big)+\bb_2\cdot \wtheta
\end{align*}
Given $\lambda \in [0,1]$, we have that 
\begin{align*}
	&\Bigg(\lambda \F^{*}\big(P_{\grad I_k}^{h_1}\big)+ (1-\lambda)\F^{*}\big(P_{\grad I_k}^{h_2}\big)(\theta)\Bigg)(\theta) \\=&\big(\lambda a_1+(1-\lambda)a_2\big)\F^{*}\big(P(\grad I_k)\big) + \big(\lambda \bb_1+(1-\lambda)\bb_2\big)\cdot \wtheta,\\
	=&  \F^{*}\big(P^{h_{\lambda}}_{\nabla I_k}\big) (\theta),
\end{align*}
where $h_{\lambda}(\bz)= \big(\lambda a_1+(1-\lambda)a_2\big)\bz +\lambda \bb_1+(1-\lambda)\bb_2 \in \mathbb H_0$.
Hence $\lambda \F^{*}\big(P_{\grad I_k}^{h_1}\big)+ (1-\lambda)\F^{*}\big(P_{\grad I_k}^{h_2}\big) = \F^{*}\big(P^{h_{\lambda}}_{\nabla I_k}\big) \in \F^{*}\big(\nSetHNot\big)$.

\subsection{Using the smallest $d^c$ in equation \eqref{eq:subspaceProj} solves the classification problem} \label{appendix:metric}
\noindent{Proposition:} Assume that for any $c\neq c^{\prime}$, there exists a $k_0$ (possibly depending on $c, c^{\prime}$) such that $	\mP_{\mH_0,{k_0}}^{c} \cap \V^{c^{\prime}}_{k_0} = \emptyset$.  Then given a test image $I^t\in \mathbb S^{c}$, 
\begin{equation}\label{eq: dc1}
	d^{c} = \sum_{k=1}^K d^{c}_k=\sum_{k=1}^k d(P_{\nabla I_k^t}, \V_k^{c})=0,
 \end{equation}
 while 
 \begin{equation}\label{eq: dc2}
	d^{c^{\prime}} = \sum_{k=1}^K d^{c^{\prime}}_k = =\sum_{k=1}^k d(P_{\nabla I_k^t}, \V_k^{c^{\prime}})>0,
 \end{equation}
 if $\V_{k_0}^{c^{\prime}}$ is a closed subspace\footnote{In practice, $\V_{k_0}^{c^{\prime}}$ is a finite dimensional space and is hence closed.}. \\
 
\noindent{Proof:}  Generally speaking, given a closed subspace $\V$ of a metric space with distance metric $d$, $d(v,\V)>0$ if and only if $v\notin \V$.  To show \eqref{eq: dc1},  it suffices  to show that $P_{\nabla I_k^t}\in \mP_{\mH_0,k}^{c}\subseteq V_k^{c}$ for $k=1,...K$, which follows from the definition of the generative model $\mathbb S^{c}$ and the fact that $I^t\in  \mathbb S^{c}$. 

On the other hand, to show \eqref{eq: dc2}, it suffices to show that there exists a $k_0$ such that $P_{\nabla I_{k_0}^t}\notin  V_k^{c^{\prime}}$, which follows from the assumption that $	\mP_{\mH_0,{k_0}}^{c} \cap \V^{c^{\prime}}_{k_0} = \emptyset$ and the fact that $P_{\nabla I_{k_0}^t}\in \mP_{\mH_0,{k_0}}^{c}$.  Hence we have that $d^{c^{\prime}}_{k_0}= d(P_{\nabla I_{k_0}^t}, \V_{k_0}^{c^{\prime}})>0$.
\subsection{Tables}
\label{sec:tables}

Table~\ref{tab:hyperParameter} provides performance comparison using different parameter configurations such as varying cell size, overlap size, and number of projections. A detailed discussion about this table is available in Section~\ref{sec-dis}.

\begin{table}[H]
\centering
\caption{}
\label{tab:hyperParameter}
\resizebox{0.48\textwidth}{!}{%
\begin{tabular}{|ccccccc|}
\hline
\multicolumn{1}{|c|}{\multirow{2}{*}{}} &
  \multicolumn{6}{c|}{Number of projections} \\ \cline{2-7} 
\multicolumn{1}{|c|}{} &
  \multicolumn{1}{c|}{2} &
  \multicolumn{1}{c|}{3} &
  \multicolumn{1}{c|}{4} &
  \multicolumn{1}{c|}{8} &
  \multicolumn{1}{c|}{20} &
  45 \\ \hline
\multicolumn{1}{|c|}{cell size = 16} &
  \multicolumn{1}{c|}{35.1$\%$} &
  \multicolumn{1}{c|}{40.0$\%$} &
  \multicolumn{1}{c|}{38.7$\%$} &
  \multicolumn{1}{c|}{39.7$\%$} &
  \multicolumn{1}{c|}{39.5$\%$} &
  40.0 \\ \hline
\multicolumn{1}{|c|}{cell size = 8} &
  \multicolumn{1}{c|}{62.4$\%$} &
  \multicolumn{1}{c|}{65.3$\%$} &
  \multicolumn{1}{c|}{63.2$\%$} &
  \multicolumn{1}{c|}{69.7$\%$} &
  \multicolumn{1}{c|}{69.4$\%$} &
  69.5 \\ \hline
\multicolumn{1}{|c|}{cell size = 4} &
  \multicolumn{1}{c|}{78.5$\%$} &
  \multicolumn{1}{c|}{86.5$\%$} &
  \multicolumn{1}{c|}{87.3$\%$} &
  \multicolumn{1}{c|}{89.3$\%$} &
  \multicolumn{1}{c|}{89.3$\%$} &
  89.3 \\ \hline
\multicolumn{7}{|c|}{Overlap size} \\ \hline
\multicolumn{1}{|c|}{} &
  \multicolumn{1}{c|}{\begin{tabular}[c]{@{}c@{}}0\\ cell size = 8\end{tabular}} &
  \multicolumn{1}{c|}{\begin{tabular}[c]{@{}c@{}}2\\ cell size = 8\end{tabular}} &
  \multicolumn{1}{c|}{\begin{tabular}[c]{@{}c@{}}4\\ cell size = 8\end{tabular}} &
  \multicolumn{1}{c|}{\begin{tabular}[c]{@{}c@{}}0\\ cell size = 4\end{tabular}} &
  \multicolumn{1}{c|}{\begin{tabular}[c]{@{}c@{}}1\\ cell size = 4\end{tabular}} &
  \begin{tabular}[c]{@{}c@{}}2\\ cell size = 4\end{tabular} \\ \hline
\multicolumn{1}{|c|}{Accuracy} &
  \multicolumn{1}{c|}{63.2$\%$} &
  \multicolumn{1}{c|}{73.2$\%$} &
  \multicolumn{1}{c|}{76.9$\%$} &
  \multicolumn{1}{c|}{87.3$\%$} &
  \multicolumn{1}{c|}{91.8$\%$} &
  93.4 \\ \hline
\end{tabular}%
}
\label{tab:hyperparameter}
\end{table}

Table~\ref{tab:log-performance} reports performance of the proposed method using log transform as a preprocessing step. A detailed discussion about this table is available in Section~\ref{sec:futurework}.

\begin{table}[H]
\caption{}
\label{tab:log-performance}
\centering
\resizebox{0.48\textwidth}{!}{%
\begin{tabular}{|c|cccc|}
\hline
 & \multicolumn{4}{c|}{Accuracy} \\ \hline
 & \multicolumn{1}{c|}{\textit{Test subset 1}} & \multicolumn{1}{c|}{\textit{Test subset 2}} & \multicolumn{1}{c|}{\textit{Test subset 3}} & \textit{Test subset 4} \\ \hline
Yale Face database (\textit{Test 1}) & \multicolumn{1}{c|}{100$\%$} & \multicolumn{1}{c|}{98.9$\%$ ($0.1\%\uparrow$)} & \multicolumn{1}{c|}{96.6$\%$ ($0.4\%\uparrow$)} & 95.7$\%$($1.3\%\uparrow$) \\ \hline
Yale Face database (\textit{Test 2}) & \multicolumn{1}{c|}{98.9$\%$($0.5\%\uparrow$)} & \multicolumn{1}{c|}{95.5$\%$} & \multicolumn{1}{c|}{95.5$\%$($3.1\%\uparrow$)} & 96.3$\%$($4.5\%\uparrow$) \\ \hline
 & \multicolumn{1}{c|}{\textit{Test subset 1}} & \multicolumn{1}{c|}{\textit{Test subset 2}} & \multicolumn{1}{c|}{\textit{Test subset 3}} & - \\ \hline
CAS-PEAL-R1 dataset & \multicolumn{1}{c|}{45.2$\%$($3.0\%\uparrow$)} & \multicolumn{1}{c|}{54.3$\%$($4.1\%\uparrow$)} & \multicolumn{1}{c|}{52.4$\%$($2.7\%\uparrow$)} & - \\ \hline
 & \multicolumn{1}{c|}{\textit{Test subset 1}} & \multicolumn{1}{c|}{\textit{Test subset 2}} & \multicolumn{1}{c|}{-} & - \\ \hline
ARFace dataset & \multicolumn{1}{c|}{99.5$\%$} & \multicolumn{1}{c|}{95.5$\%$($\uparrow 1.0\%$)} & \multicolumn{1}{c|}{-} & - \\ \hline
\end{tabular}
}
\end{table}
\end{document}